%% file: neurips_2024.tex
\documentclass{article}

\usepackage[nonatbib, final]{neurips_2024}

\usepackage[utf8]{inputenc} %
\usepackage[T1]{fontenc}    %
\usepackage{hyperref}       %
\usepackage{url}            %
\usepackage{booktabs}       %
\usepackage{amsfonts}       %
\usepackage{nicefrac}       %
\usepackage{microtype}      %
\usepackage{xcolor}         %
\usepackage{soul}
\usepackage{tcolorbox}
\usepackage[numbers]{natbib}
\usepackage{geometry}
\usepackage{amsmath, bm}
\usepackage{amssymb}
\usepackage{listings}
\usepackage{mathtools}
\usepackage{siunitx}
\usepackage{xspace}
\usepackage{ntheorem}
\usepackage{graphicx}
\usepackage{booktabs}
\usepackage{bbm}
\usepackage{stmaryrd}
\usepackage{algorithm}
\usepackage{subcaption}
\usepackage{wrapfig}
\usepackage{algorithmic}

\input{commands}

\title{Leveraging Environment Interaction for Automated PDDL Translation and Planning with Large Language Models}

\author{%
  Sadegh Mahdavi\thanks{Work performed while interning at Borealis AI.} \\
  University of British Columbia\\
  \texttt{smahdavi@ece.ubc.ca} \\
  \And
  Raquel Aoki \\
  Borealis AI \\
  \texttt{raquel.aoki@borealisai.com} \\
  \AND
  Keyi Tang \\
  Borealis AI \\
  \texttt{keyi.tang@borealisai.com} \\
  \And
  Yanshuai Cao \\
  Borealis AI \\
  \texttt{yanshuai.cao@borealisai.com} \\
}

\begin{document}

\maketitle

\begin{abstract}
Large Language Models (LLMs) have shown remarkable performance in various natural language tasks, but they often struggle with planning problems that require structured reasoning. To address this limitation, the conversion of planning problems into the Planning Domain Definition Language (PDDL) has been proposed as a potential solution, enabling the use of automated planners. However, generating accurate PDDL files typically demands human inputs or correction, which can be time-consuming and costly. In this paper, we propose a novel approach that leverages LLMs and environment feedback to automatically generate PDDL domain and problem description files without the need for human intervention. Our method introduces an iterative refinement process that generates multiple problem PDDL candidates and progressively refines the domain PDDL based on feedback obtained from interacting with the environment. To guide the refinement process, we develop an Exploration Walk (EW) metric, which provides rich feedback signals for LLMs to update the PDDL file. We evaluate our approach on $10$ PDDL environments. We achieve an average task solve rate of 66\% compared to a 29\% solve rate by GPT-4's intrinsic planning with chain-of-thought prompting. Our work enables the automated modeling of planning environments using LLMs and environment feedback, eliminating the need for human intervention in the PDDL translation process and paving the way for more reliable LLM agents in challenging problems. Our code is available at \hyperref[https://github.com/BorealisAI/llm-pddl-planning]{https://github.com/BorealisAI/llm-pddl-planning}
\end{abstract}

\input{secs/0_intro}
\input{secs/1_related_work}
\input{secs/2_background}
\input{secs/3_method}

\input{secs/4_experiments}
\input{secs/5_conclusion}

\bibliographystyle{abbrvnat}
\bibliography{refs}

\newpage
\appendix
\input{secs/appendix}

\newpage
\section*{NeurIPS Paper Checklist}

\begin{enumerate}

\item {\bf Claims}
    \item[] Question: Do the main claims made in the abstract and introduction accurately reflect the paper's contributions and scope?
    \item[] Answer: \answerYes{} %
    \item[] Justification: The claims are supported by experiments.

\item {\bf Limitations}
    \item[] Question: Does the paper discuss the limitations of the work performed by the authors?
    \item[] Answer: \answerYes{} %
    \item[] Justification: In the conclusion section, we discuss limitations of our work.

\item {\bf Theory Assumptions and Proofs}
    \item[] Question: For each theoretical result, does the paper provide the full set of assumptions and a complete (and correct) proof?
    \item[] Answer: \answerNA{} %
    \item[] Justification: Our paper does not include theoretical results.

    \item {\bf Experimental Result Reproducibility}
    \item[] Question: Does the paper fully disclose all the information needed to reproduce the main experimental results of the paper to the extent that it affects the main claims and/or conclusions of the paper (regardless of whether the code and data are provided or not)?
    \item[] Answer: \answerYes{} %
    \item[] Justification: We describe our method in Algorithm \ref{alg:method}, and Appendix \ref{apx:method_details}. The code will be made publicly available upon publication.

\item {\bf Open access to data and code}
    \item[] Question: Does the paper provide open access to the data and code, with sufficient instructions to faithfully reproduce the main experimental results, as described in supplemental material?
    \item[] Answer: \answerNo{} %
    \item[] Justification:  The code will be made publicly available upon publication. We explain our method in Algorithm \ref{alg:method}. For the data, the PDDL files are publicly available, and we provide examples on how to obtain natural language descriptions.

\item {\bf Experimental Setting/Details}
    \item[] Question: Does the paper specify all the training and test details (e.g., data splits, hyperparameters, how they were chosen, type of optimizer, etc.) necessary to understand the results?
    \item[] Answer: \answerYes{} %
    \item[] Justification: Our method uses pre-trained large language models, and we provide examples in Algorithm \ref{alg:method} on how we prompt the LLMs.

\item {\bf Experiment Statistical Significance}
    \item[] Question: Does the paper report error bars suitably and correctly defined or other appropriate information about the statistical significance of the experiments?
    \item[] Answer: \answerYes{} %
    \item[] Justification: Figure \ref{fig:ew_corr} and Figure \ref{fig:ew_corr_full} show standard error. Table \ref{tab:final_table} does not contain any confidence intervals as the experiments become computationally expensive.

\item {\bf Experiments Compute Resources}
    \item[] Question: For each experiment, does the paper provide sufficient information on the computer resources (type of compute workers, memory, time of execution) needed to reproduce the experiments?
    \item[] Answer: \answerYes{} %
    \item[] Justification: In the experiments section, we reported the number of tokens used to create Table \ref{tab:final_table}.
    
\item {\bf Code Of Ethics}
    \item[] Question: Does the research conducted in the paper conform, in every respect, with the NeurIPS Code of Ethics \url{https://neurips.cc/public/EthicsGuidelines}?
    \item[] Answer: \answerYes{} %
    \item[] Justification: Yes

\item {\bf Broader Impacts}
    \item[] Question: Does the paper discuss both potential positive societal impacts and negative societal impacts of the work performed?
    \item[] Answer: \answerNA{} %
    \item[] Justification: The setup used in our work does not have direct societal impact as it is centered around generating PDDL code.
    
\item {\bf Safeguards}
    \item[] Question: Does the paper describe safeguards that have been put in place for responsible release of data or models that have a high risk for misuse (e.g., pretrained language models, image generators, or scraped datasets)?
    \item[] Answer: \answerNA{} %
    \item[] Justification: Our work is a fundamental work and poses no such risk.

\item {\bf Licenses for existing assets}
    \item[] Question: Are the creators or original owners of assets (e.g., code, data, models), used in the paper, properly credited and are the license and terms of use explicitly mentioned and properly respected?
    \item[] Answer: \answerYes{} %
    \item[] Justification: We have cited the sources from which we use library, data, and code.

\item {\bf New Assets}
    \item[] Question: Are new assets introduced in the paper well documented and is the documentation provided alongside the assets?
    \item[] Answer: \answerNA{} %
    \item[] Justification: The code will be made publicly available upon publication 

\item {\bf Crowdsourcing and Research with Human Subjects}
    \item[] Question: For crowdsourcing experiments and research with human subjects, does the paper include the full text of instructions given to participants and screenshots, if applicable, as well as details about compensation (if any)? 
    \item[] Answer: \answerNA{} %
    \item[] Justification: The paper does not involve crowdsourcing nor research with human subjects.

\item {\bf Institutional Review Board (IRB) Approvals or Equivalent for Research with Human Subjects}
    \item[] Question: Does the paper describe potential risks incurred by study participants, whether such risks were disclosed to the subjects, and whether Institutional Review Board (IRB) approvals (or an equivalent approval/review based on the requirements of your country or institution) were obtained?
    \item[] Answer: \answerNA{} %
    \item[] Justification: The paper does not involve crowdsourcing nor research with human subjects. 

\end{enumerate}

\end{document}

%% file: commands.tex
\DeclareMathOperator*{\argmax}{argmax} %

\newcommand{\new}[1]{{#1}}

\newcommand{\mdhat}{m_{d \leftrightarrow \hat{d}}}
\newcommand{\llm}{\text{LLM}\xspace}
\newcommand{\nl}{\text{NL}\xspace}

\newcommand{\Pcal}{\mathcal{P}}
\newcommand{\Dcal}{\mathcal{D}}
\newcommand{\Acal}{\mathcal{A}}

\newcommand{\Scal}{\mathcal{S}}

\newcommand{\Ebb}{\mathbb{E}}
\newcommand{\Pbb}{\mathbb{P}}
\newcommand{\1}{\mathbbm{1}}

\newcommand{\planerr}{\bot}

\newcommand{\ie}{\textit{i.e.}}
\newcommand{\eg}{\textit{e.g.}}

\newtheorem{definition}{Definition}

\newtheorem{assumption}{Assumption}

\definecolor{highlight_blue}{HTML}{bfefff}
\definecolor{highlight_yellow}{HTML}{f0ff6e}
\newcommand{\highlb}[1]{\sethlcolor{highlight_blue}\hl{#1}}
\newcommand{\highly}[1]{\sethlcolor{highlight_yellow}\hl{#1}}

\lstdefinelanguage{pddl}{
    basicstyle=\ttfamily\fontsize{8}{10}\selectfont,
    sensitive=true,
    morecomment=[l]{;}, 
    morestring=[b]",
    keywords={define,domain,problem,agent,object,requirements,types,predicates,action,parameters,precondition,postcondition,goal,init,forall,exists,and,or,not,imply,when,either,effect},
    keywordstyle=\color{blue!70!black}\bfseries,
    morekeywords={[2]},
    keywordstyle={[2]\color{green!40!black}\bfseries},
    morekeywords={[3]},
    keywordstyle={[3]\color{orange!80!black}\bfseries},
    basicstyle=\ttfamily\fontsize{10}{12}\selectfont,
    commentstyle=\color{gray}\itshape,
    stringstyle=\color{red!70!black},
    showstringspaces=false,
    numbers=left,
    numberstyle=\tiny\color{gray},
    numbersep=8pt,
    tabsize=2,
    breaklines=true,
    breakatwhitespace=false,
    backgroundcolor=\color{gray!10},
    captionpos=b,
    belowcaptionskip=0.5em,
    escapeinside={(*@}{@*)},
}

\lstdefinelanguage{markdown}{
    basicstyle=\ttfamily\fontsize{8}{10}\selectfont,
    sensitive=false,
    morecomment=[l]{\#},
    morecomment=[s]{<!--}{-->},
    morestring=[b]",
    morestring=[b]',
    morestring=[s]{``}{''},
    morestring=[s]{`}{`},
    keywordstyle=\color{blue!70!black}\bfseries,
    commentstyle=\color{green!50!black}\itshape,
    stringstyle=\color{purple!70!black},
    basicstyle=\ttfamily\fontsize{10}{12}\selectfont,
    showstringspaces=false,
    numbers=left,
    numberstyle=\tiny\color{gray},
    numbersep=8pt,
    tabsize=2,
    breaklines=true,
    breakatwhitespace=true,
    backgroundcolor=\color{gray!10},
    captionpos=b,
    belowcaptionskip=0.5em,
    escapeinside={(*@}{@*)},
}

%% file: secs/0_intro.tex
\section{Introduction}

Large language models (LLMs) have demonstrated remarkable success across various domains, including mathematics, coding, and even the bar exam \cite{gpt4}. These models excel at understanding and generating natural language, offering flexibility and adaptability to a wide range of tasks. However, when it comes to planning and long-horizon reasoning, LLMs have shown limited performance \cite{dziri2024faith, valmeekam2023planbench}, despite some promising results \cite{bubeck2023sparks}.

Planning is a crucial aspect of intelligence that involves reasoning to find a sequence of actions to achieve a desired goal state from an initial state. The Planning Domain Definition Language (PDDL) \cite{pddl} is a widely used formalism for describing planning problems. PDDL provides a structured way to define the domain, which includes the types of objects, predicates, and actions, as well as the problem instance, which specifies the initial state and goal conditions. PDDL enables the application of search-based algorithms, such as breadth-first search (BFS) or A\textsuperscript{$*$} search, which can guarantee to find a valid solution if one exists. However, the downside of PDDL is that it requires a well-defined and structured domain and problem definition, which can be challenging to create, especially for complex scenarios. Figure \ref{fig:grippers_pddl_example} showcases snippets of some PDDL problems and domain files along with an action plan produced by a classical planner.

Recent studies explored combining the strengths of LLMs and PDDL-based planning~\cite{LLMP, LLM-DP, guan2023leveraging}. The idea is to leverage LLM for translation from natural language (NL) problem descriptions into PDDL formal descriptions, and then use a classical planner to solve the translated PDDL problem \cite{guan2023leveraging}. This hybrid approach could theoretically take advantage of the flexibility of NL input and the correctness guarantees provided by the classical planner. If the translation from NL to PDDL is accurate, the resulting plan is guaranteed to be valid. 

Unfortunately, existing approaches have not been able to generate both PDDL problem and domain descriptions with reasonable success rates without humans in the loop, as we shall elaborate in Sec.\ \ref{sec:related}. While translating PDDL problems is feasible given the domain PDDL description \cite{LLMP}, generating domain PDDL from NL correctly is a more nuanced and challenging problem. 
To do so requires identifying causally relevant objects to design predicates, as well as their inter-relationships, in a way that accurately reflects the possible states and transitions of the environment. A small error, for example in predicate design, could lead to entirely incorrect domain description and failed planning (see Appendix \ref{apx:predicate_design} for a real example). \citet{guan2023leveraging} take a step toward this goal relying on human-in-the-loop to detect and correct mistakes made by LLMs. 

In this work, we develop a fully automated method for generating PDDL domain and problem definitions using LLMs and environment feedback without relying on human intervention. Intuitively, our method lets an LLM build hypothetical ``mental models'' of the environment, in the form of proposed PDDL domain descriptions. The LLM then verifies and updates the ``mental model'' by observing discrepancies between the feasibility of actions under its ``mental model'' and the real environment. This method enables LLMs to use classical planners to solve complex planning problems whose solutions may require hundreds or thousands of steps that all need to be correct. 

We first highlight the challenges of this task and then propose our solution. In particular, our contributions are as follows:
\begin{itemize}
\item We demonstrate that even small modifications to PDDL domains can render plan search infeasible, limiting the feedback information for LLMs to perform in context update. 

\item To address this, we introduce a new Exploration Walk (EW) metric, which is a smooth similarity measure between two domains by comparing the executability of random action sequences sampled from one domain on the other. Crucially, EW only requires access to the action interface and executability of the environments, not directly the ground-truth PDDL.

\item We propose an EW-guided tree search approach that leverages LLMs to generate and refine the PDDL domain and problem files iteratively and automatically. 
\item We evaluate our method on $10$ challenging PDDL domains, where a number of them are from the International Planning Competition, and show that it outperforms a baseline that generates PDDL files in a single attempt without refinement. Our method solves 7 out of 10 environments, achieving an average task solve rate of 66\% and average EW score of 0.84, compared to 34\% task solve rate and 0.53 EW score for the baseline, and 29\% solve rate by GPT-4 (\texttt{gpt-4-1106-preview})'s intrinsic planning with chain-of-thought prompting.
   
\end{itemize}

To the best of our knowledge, this is the first work that enables modeling a planning environment via PDDL translation using LLMs and environment interaction, without the need for human intervention. 

%% file: secs/1_related_work.tex
\section{Related Work}
\label{sec:related}

\noindent\textbf{LLMs and Classical Planning.} There has been recent interest in integrating LLMs with PDDL~\cite{LLMP, valmeekam2023planbench, guan2023leveraging, LLM-DP, xie2023translating, silver2022pddl, han2024interpret, oswald2024large, stein2023autoplanbench}, and more generally neural networks with PDDL~\cite{silver2023predicate, ahmetoglu2022deepsym}. \citet{silver2024generalized} leverage LLMs to take domain PDDLs and problem PDDL specifications, and synthesize a Python function to generate domain-specific plans, as a replacement for search-based planning. 
\citet{LLMP} show that using LLMs to translate problem specification to PDDL, and using classical solvers results into a higher planning accuracy that using LLM directly as a planner. \citet{LLM-DP} consider a similar setting, but assume that the list of objects is partially observable, and the LLM needs to interact with the world to observe the list of objects. All of the mentioned works, however, assume that a domain PDDL files is already provided. 
\new{\citet{oswald2024large} generate domain PDDL from natural language and propose heuristics for comparing PDDL action domains. However, their approach assumes that predicates are provided, whereas our work makes no such assumption. Additionally, \citet{oswald2024large} rely on ground-truth problem instances for domain compatibility evaluation, whereas we directly translate problem PDDL without any such assumptions.}
\citet{guan2023leveraging} translate both Domain and Problem from natural language description but rely on human experts to correct mistakes in the domain translation before generating problem PDDLs. In this work, our goal is to lift the human-intervention assumption, and instead, use domain interaction for evaluation and verification. See Table \ref{tab:related_work} for a summary of related work comparison.

\noindent\textbf{\new{Direct} Reasoning with LLMs.} Recent research has explored eliciting \new{direct} reasoning capabilities within Large Language Models (LLMs). This reasoning can be either entirely \new{direct}~\citep{yao2023tree, wei2022chain} or partially \new{direct} with the assistance of basic external tools~\cite{faithful_cot}. However, the primary limitation of these approaches lies in the inherent tendency of auto-regressive LLMs to produce errors in long-horizon reasoning tasks~\cite{valmeekam2023planbench}. Even a minor mistake in a single reasoning step can lead to cascading errors, ultimately resulting in an incorrect final answer~\cite{dziri2024faith}. When applied to classical planning, this approach delegates the entire plan generation process to an LLM instead of leveraging a dedicated classical planner. Studies have demonstrated that this strategy is suboptimal compared to generating PDDL code directly~\cite{guan2023leveraging, LLMP}, highlighting the importance of incorporating classical planning tools for faithful plan generation in classical planning tasks.

\noindent\textbf{External Reasoning and Code Generation.} This last line of work focuses on generating executable code from natural language instructions such as SQL or Python code generation~\cite{codex, ni2023lever, self_refine, self-debug, faithful_cot, zhang2023coder}. Here, the LLM often acts as a code translator, and the reasoning logic lies within the generated code. \citet{codex} show that LLMs are capable of Python code generation from docstrings to high accuracy. The authors also find that taking multiple code samples from an LLM and picking the best samples results in an accuracy boost. Later works show that iterative refinement of LLM responses improves the accuracy on the downstream task~\cite{self_refine, self-debug}, especially given external feedback such as unit tests or human feedback. Our work is related to code generation as we produce structured PDDL files. However, our setting presents three challenges: (1) there are two types of PDDL files, in contrast to a single Python script, and the two files need to be consistent with each other; (2) more importantly, getting external feedback and the evaluation of a generated PDDL code is not as easy as python unit tests, and as we show in Section \ref{sec:ew_intro}, (domain generation) errors are abundant and hard to trace; (3) LLMs are trained with a lot more Python code compared to PDDL, as the later is much scarcer. 

\begin{table}[t]
\caption{Summary of comparison to most closely related prior studies.$^{\star}$Require at least one problem instance to be translated by a human into the target domain as an in-context example.}
\label{tab:related_work}
\centering
\resizebox{.8\textwidth}{!}{%
\begin{tabular}{lccc}
\toprule
\textbf{Method(s)} & \textbf{Translate Problem} & \textbf{Translate Domain} & \textbf{No Human Intervention} \\ \midrule
LLM+P~\citep{LLMP}, LLM-DP~\citep{LLM-DP} & $\checkmark^\star$  & $\times$ & \checkmark  \\ \midrule
LLM World Models~\citep{guan2023leveraging} & \checkmark  & \checkmark   & $\times$  \\ \midrule
Ours  & \checkmark   & \checkmark   & \checkmark \\
\bottomrule
\end{tabular}
}
\end{table}

%% file: secs/2_background.tex
\section{Notation and Background}
\noindent \textbf{Notation.} We denote $\1[\cdot]$ as the indicator function. The notation $1:N$ refers to the sequence of integers ranging from $1$ to $N$. For a set $\Acal$, we define $\Acal^*$ as the set comprising all possible sequences of elements drawn from $\Acal$, and define $2^\Acal$ as the power set of $\Acal$.

\noindent\textbf{PDDL.} Planning Domain Definition Language (PDDL) is a formal language used to describe and specify planning problems for automated planning. Here, we have two types of PDDL files: (1) \emph{Domain PDDL}, which defines possible \emph{predicates} (\ie, states), and \emph{actions} in the environment. Executing each action requires some \emph{precondition} (\ie, a set of predicates to have a specific value), and the execution leads to some \emph{effect} (\ie, a change in the values of some predicates). (2) \emph{Problem PDDL}, which contains a set of initial predicates and a set of goal predicates. 

The problem PDDL instantiates the domain definition PDDL to form a concrete environment. Together,  the planning problem is fully defined and formalized. 
A \emph{classical planner} takes in both files and searches for a plan based on the provided specification. 
A \emph{plan} is a sequence of actions, starting from the initial state, leading to a state satisfying the goal conditions, with each action respecting the rules of the environment. 
Formally, let $\Dcal, \Pcal, \Acal$ be the set of all possible domains, problems, and actions, respectively. 
Then, given a domain $d \in \Dcal$ and problem $p \in \Pcal$, a classical planner $C: \Dcal \times \Pcal \rightarrow \Acal^*\cup \{\planerr\}$ takes in domain $d$ and plan $p$, and produces a plan $q := C(d, p)$ which is either set of actions from $\Acal^*$, or a planning error $\planerr$.
A planning error may be due to an infeasible plan search (\ie, plan not found), syntax errors, or incompatible domain and problem. A plan validator verifies whether a plan $q$ is executable and achieves the desired problem goal given a domain PDDL $d$ and problem PDDL $p$, \ie,  whether $q$ solves the planning problem instance. The validator function, denoted as  $V_{d, p}(q):  \Acal^* \rightarrow \{0, 1\}$, is $1$ if the plan is valid, and $0$ otherwise. For convenience, we assume $V_{d, p}(\planerr) = 0$. Similarly, we define plan execution checker $E_{d, p}: \Acal^* \rightarrow \{0, 1\}$, which only checks whether an action sequence is executable in a domain or not. Note that the \emph{difference between $V$ and $E$ is that the former checks for both plan executability and goal satisfaction, while the latter only checks for plan executability.}
We also define $\Scal$ as the set of all possible states. Function $A_{d, p}: \Scal \rightarrow 2^\Acal$ delineates the set of legal actions given the current states (\ie, actions that would not immediately result in $E_{d, p}$ returning 0). The function $S_{d, p}: \Acal \times \Scal \rightarrow \Scal$ denotes the state transition function (\,  i.e., $S_{d, p}(a, s)$ determines the subsequent state given the current state $s$ and action $a$). Finally, we denote the initial state induced by $d$ and $p$ to be $s_{d, p, 0} \in \Scal$. See Table \ref{tab:notation} in the Appendix for a summary of notations.

To illustrate the definitions with an example, consider the Grippers~\cite{ipc_comp} environment with several rooms containing robots and boxes. Robots can move balls between rooms using their left and right grippers. Given an initial setting of robots and balls in different rooms, the main goal is to move specific balls to specific rooms using the robots.
\begin{wrapfigure}{r}{0.6\textwidth}
    \centering
    \includegraphics[width=0.59\textwidth]{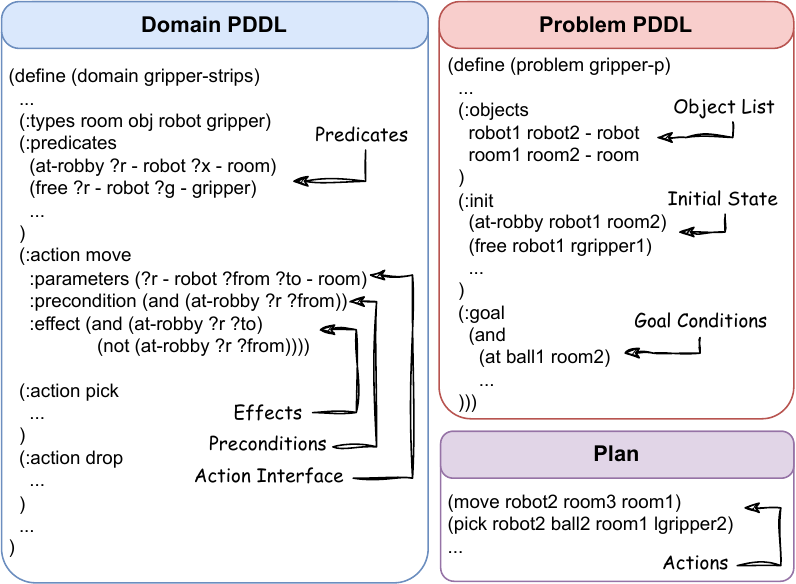}
    \caption{Snippets of PDDL domain, problem, and plan.}
    \label{fig:grippers_pddl_example}
\end{wrapfigure}
Figure \ref{fig:grippers_pddl_example} shows an annotated example domain, problem, and plan for this environment. The domain determines predicates and actions. Predicates such as \texttt{at-robby} keep track of object states (\eg, whether a particular robot is in a particular room) and defining suitable predicates is a crucial part of domain design. The \texttt{move} action for moving a robot from one room to another has three parameters: robot \texttt{r}, departure room \texttt{from}, and destination room \texttt{to}. Each action has preconditions and effects, which comprise the main logic of the domain for determining the actionability of an action. In the case of the \texttt{move} action, the precondition is that the robot must be in the \texttt{from} room, and the effect is that it will no longer be in that room and will be in the \texttt{to} room. A problem PDDL $p$ specifies the initial state of robots, boxes, rooms, and the final goal. For instance, \texttt{(at-robby robot2 room3)} means that \texttt{robot2} is initially at \texttt{room3}. The predicate \texttt{(at ball1 room2)} specifies the goal condition that \texttt{ball1} must eventually be moved to \texttt{room2}. A plan constitutes a sequence of actions to reach the goal. For instance, one action could be \texttt{(move robot2 room3 room1)}, moving \texttt{robot2} from \texttt{room3} to \texttt{room1}. If \texttt{robot2} is not already in \texttt{room3}, this action is considered illegal, and the environment will produce an error. For a complete example of domain $d$, problem $p$, and plan $q$, see Listings \ref{lst:grippers_domain_pddl}, \ref{lst:grippers_problem_pddl}, and \ref{lst:grippers_problem_plan}, respectively in the Appendix.

\noindent\textbf{Large Language Models (LLMs).} We assume access to a powerful language model $\llm$. $\llm_n(X)$ denotes sampling $n$ responses from the $\llm$ given prompt $X$. 
Following the prior works, we set a temperature of $\tau=0$ for sampling with $n=1$ (\ie, greedy sampling), and a temperature of $\tau=0.7$ for $n>1$~\cite{self-debug}. Whenever possible, we use zero-shot or one-shot chain-of-thought prompts~\cite{zero_cot, wei2022chain} for the $\llm$ to reason before generating a response.

%% file: secs/3_method.tex
\section{Method}
Given an environment $e$, its domain NL description and a task NL description, the environment's object list and action interface, our goal is to model the environment by generating a domain PDDL $\hat{d} \in \Dcal$ and a problem PDDL $\hat{p} \in \Pcal$, such that applying a classical planner $C$ on the PDDL files produces a valid plan for the environment, \ie, $C(\hat{d}, \hat{p})$ is a valid plan for $e$, \ie, $V_{d,p}(C(\hat{d}, \hat{p})) = 1$.

\subsection{Setup}

For evaluation, we assume there exists a ground truth domain PDDL $d \in \Dcal$, and a corresponding problem instance $p \in \Pcal$. However, the ground truth is not directly compared to generated $\hat{d}, \hat{p}$, but to validate the plan $\hat{q} := C(\hat{d}, \hat{p})$ by executing the validator of the ground-truth environment, $V_{d, p}(\hat{q})$. 

Formally, for each environment $e$ with domain PDDL $d \in \Dcal$, and $N$ tasks with their corresponding ground-truth problem PDDLs $p_{1:N} := (p_1, p_2, \dots, p_N), p_{1:N} \in \Pcal^N$, our goal is to generate a domain PDDL $\hat{d}$, and a sequence of task PDDLs $\hat{p}_{1:N} := (\hat{p}_1, \hat{p}_2, \dots, \hat{p}_N)$ such that the average solve rate $\overline{V}$ is maximized: 
\begin{equation}
\argmax_{\hat{d} \in \Dcal, \hat{p}_{1:N} \in \Pcal^N} \overline{V}(\hat{d}, \hat{p}_{1:N} ; e) := \frac{1}{N} \sum_{i=1}^{N} V_{d, p_i} \left(C(\hat{d}, \hat{p}_i) \right). \label{eq:tasks_solved}
\end{equation}

Generating accurate $\hat{d}$ and $\hat{p}_{1:N}$ in one attempt is often impractical \cite{guan2023leveraging}, and some form of feedback is required to refine the response. \citet{guan2023leveraging} leverage human expert feedback on $\hat{d}$ to correct the generated domain. However, human feedback may not always be reliable and is not scalable. Before introducing our method that relies on environment feedback instead, we first state our assumptions:
\begin{assumption}[Environment access] \label{asm:env} We assume the list of objects and action interfaces are known. Furthermore, we assume that executability and verifiability of actions can be observed (through the functions $E_{d, p}$ and $V_{d, p}$).
\end{assumption}
\begin{assumption}[Natural language description] \label{asm:description}
We assume the natural language descriptions of the domain and task are both given.
\end{assumption}

The action interfaces are equivalent to APIs available to LLM agents. So it is reasonable to assume that the exact API call signatures are known. On the other hand, one may wonder why the object list, which appears in problem PDDLs as illustrated in Figure \ref{fig:grippers_pddl_example} needs to be assumed to be given, when the NL problem description should describe the objects involved in the planning tasks. 
This is because the NL description may not refer to the object instances using exactly the same label as the environment induced by $d$ and $p$. If $p$ refers to a robot as \texttt{robot1} but the user specifying the natural language problem description calls it \texttt{Jarvis}, then the environment only recognizes \texttt{robot1} and not \texttt{Jarvis}, so the LLM would have no way to correct this mistake due to trivial name mismatch.
See Appendix \ref{apx:dataset_details} for a detailed example of our assumptions on the Grippers environment. 

Note that our assumptions do not require the underlying environment to be a PDDL environment, but it can be any environment as long as PDDL is expressive enough to capture the working mechanisms of the environment. For digital agents in virtual environments, the list of objects and action interfaces are just different data objects and APIs available. The assumptions could even hold true for physical agents in the real world, provided recognition and control are sufficiently accurate. In this work, we focus on PDDL environments only, although our framework is more general.

\subsection{Difficulty of domain PDDL generation} \label{sec:d_challenges}

Generating the correct domain PDDL is challenging, as small mistakes could make the plan search fail. To demonstrate this brittleness, we simulate random omission of $k$ terms, where $0 \leq k \leq 10$, from the action precondition and effects of the original domain $d$. For instance, in the case of the Grippers (Figure \ref{fig:grippers_pddl_example}), we may create a new synthetic domain by removing the \texttt{(at robby ?r ?to)} term from the effects of the \texttt{move} action. Namely, we define $\hat{d}_k \sim \Pbb_k(d)$, where $\Pbb_k(d)$ represents the uniform random removal of $k$ terms. Then, for each generated $\hat{d}_k$, coupled with the ground truth task PDDLs, we compute whether the classical planner is able to find a plan without error and compute the \emph{Plan-Not-Found} rate under $k$ omissions, $\text{PNF}_k$, of the environment.

\begin{figure}
    \centering
    \begin{subfigure}{0.43\textwidth}
        \centering
        \includegraphics[width=\textwidth]{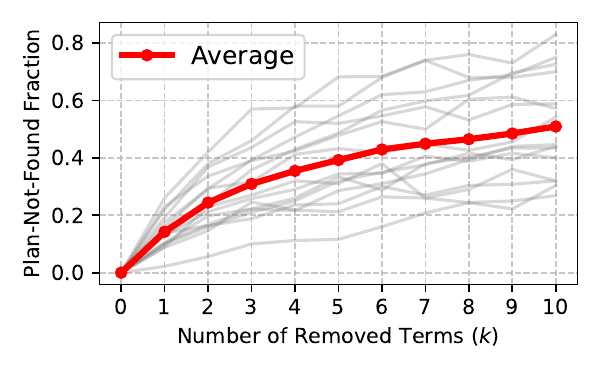}
        \caption{}
        \label{fig:pnf_metric}
    \end{subfigure}
    \hfill
    \begin{subfigure}{0.56\textwidth}
        \centering
        \includegraphics[width=\textwidth, trim=0 12.5cm 8.3cm 0, clip]{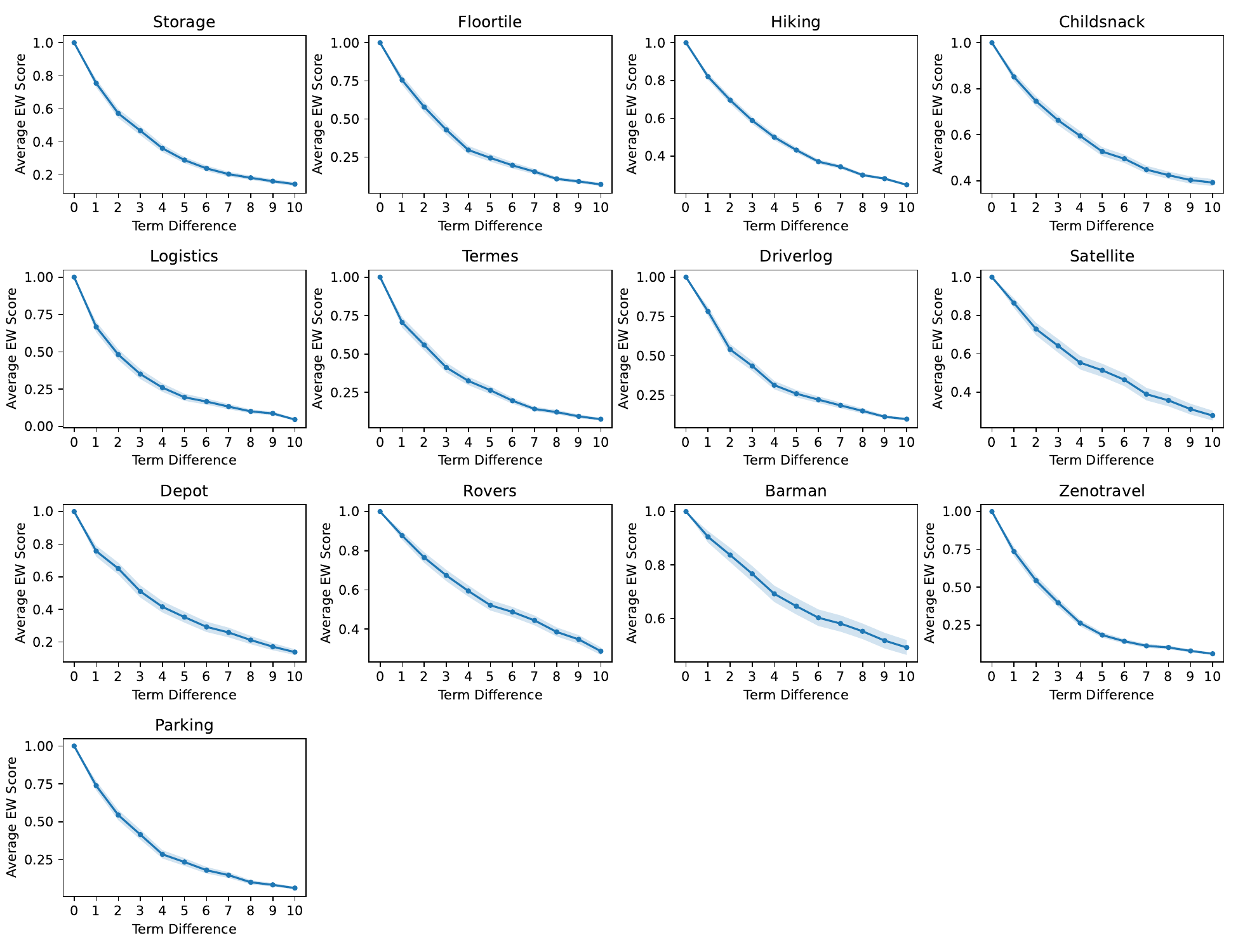}
        \caption{}
        \label{fig:ew_corr}
    \end{subfigure}
    \caption{(a) Effect of the number of removed terms on plan search failure. Each gray line shows the $\text{PNF}_k$ (Plan-Not-Found) metric for one environment. The red line is the average of all 15 environments. (b) Correlation between average exploration walk (EW) score and average domain difference. The $x$-axis shows how many terms each pair of domains differs in. The $y$-axis shows the average EW score over various pairs. All the domains show the average monotonicity of the EW score with respect to term difference.}
    \label{fig:combined}
\end{figure}

We empirically measure the value of $\text{PNF}_k$ using Monte-Carlo estimation on $15$ environments. As shown in Figure \ref{fig:pnf_metric}, $\text{PNF}_1$ has an average of $0.14$ among different environments. This means that on average $14\%$ of the terms in domain PDDLs are so critical that removing them results in a plan-not-found error. This situation is exacerbated for larger $k$: at $k=3$, the average $\text{PNF}_k$ reaches around $0.3$. In practice, the problem PDDL $\hat{p}_i$ also needs to be generated, and the generated domain $\hat{d}$ may have extra terms, both of which may further increase the planning-not-found rate.

\subsection{Domain alignment measure via Exploration Walk metrics}
\label{sec:ew_intro}

Whenever the plan search fails, absolutely no information is available to the LLM about which part of the problem or domain has issues. This is because the underlying search algorithm (such as BFS and A\textsuperscript{$*$}) fails and as a result, it does not produce any output. For example, with BFS, it enumerates all paths (possibly several thousand paths or more), and finds none satisfy the goal conditions, leaving the plan search without any useful insights. As an alternative, we introduce the Exploration Walk (EW): a smooth feedback signal that provides incremental feedback for LLM in-context learning. EW both provides a mechanism to gather richer feedback information that feeds into LLM context for PDDL refinement, as well as computing a smooth scoring metric that to compare multiple PDDLs and guide the refinement process forward.

Intuitively, the idea is to take legal random action sequences and verify their executability under LLM's "mental model" environment induced by an LLM-generated PDDL domain. This is analogous to the \emph{retrodiction} step in scientific methodology, where existing observations and experimental data need to be explained by the existing model.

And in the other direction, EW takes executable random action sequences from an LLM-generated PDDL domain and verifies whether they are correct in the real environment. This is analogous to \emph{hypothesis testing} in scientific methodology, where new predictions are verified experimentally.

We now describe the EW and EW metrics formally.
We define an Exploration Walk of length $T$ to be any action sequence sampled from a strictly positive distribution $\Pbb_{d, p, T}$ over executable $T$-step action sequences in $\Acal^*$ corresponding to domain $d$ and task $p$. We assume the probability of non-executable action sequences to be zero under $\Pbb_{d, p, T}$.  In other words, $\forall q_{1:T} \text{,~} \Pbb_{d, p, T}(q_{1:T}) > 0 \text{~iff~} E_{d,p}(q_{1:T})=1$.

For the rest of this paper, we use the simplest possible EW, with a uniform distribution over valid actions at each step. Note that to sample uniform random EW from the ground truth environment induced by $d$ and $p$, we do not need direct access to the full $d$ and $p$. We only need the list of objects in $p$ and the action interface in $d$, and executability checker $E_{d,p}$, consistent with our Assumption \ref{asm:env}. At each step, running $E_{d,p}$ on all possible actions yields the legal actions at that step for EW.

Given an EW distribution, we define an EW metric using the fractions of executability of EW walks from one domain under another, averaged over all different lengths.

\begin{definition}[EW Metrics]
Let $p_{1:N}$ and $\hat{p}_{1:N}$ be problems in domain $d$ and $\hat{d}$ respectively, such that the set of objects in $p_j$ and $\hat{p}_j$ are consistent. We define the one-sided measure $m_{d \shortrightarrow \hat{d}}$ and the symmetric one $m_{d \leftrightarrow \hat{d}}$ for the degree of alignment between two domains $d$ and $\hat{d}$ as:
\begin{align}
   m_{d \shortrightarrow \hat{d}}(p_{1:N}, \hat{p}_{1:N}) &:= \frac{1}{NT_\text{max}} \sum\nolimits_{j=1}^{N} \sum\nolimits_{T=1}^{T_\text{max}}\Ebb_{q \sim \Pbb_{d, p_j, T}} \left[ E_{\hat{d}, \hat{p}_j}(q) \right] \nonumber \\
   m_{d \leftrightarrow \hat{d}}(p_{1:N}, \hat{p}_{1:N}) &:= 2  \big/ {\left( 1 / m_{d \shortrightarrow \hat{d}}(p_{1:N}, \hat{p}_{1:N}) + 1/m_{\hat{d} \shortrightarrow d}(\hat{p}_{1:N}, p_{1:N}) \right)}, \label{eq:ew_metric}
\end{align}
where $T_\text{max}$ is the largest EW walk length. 
\end{definition}

$m_{d \shortrightarrow \hat{d}}$ measures what fraction of EWs sampled from domain $d$ are executable on the domain $\hat{d}$. Then, $\mdhat$ takes the harmonic mean of $m_{d \shortrightarrow \hat{d}}$ and $m_{\hat{d} \shortrightarrow d}$ to produce the final EW measure. This metric has two favourable properties: (1) it ensures that $m_{d \leftrightarrow \hat{d}} = m_{\hat{d} \leftrightarrow d}$, thereby providing a consistent measure of similarity regardless of the order of domain comparison. (2) the harmonic mean is resistant to trivial domain similarity inflation. By employing the harmonic mean rather than the arithmetic mean, the symmetric EW metric prevents domains that are overly permissive (e.g., domains where all actions are permissible without any preconditions) from being similar to more restrictive domains. For example, in a scenario where domain $\hat{d}$ allows all possible actions without restrictions, $m_{d \shortrightarrow \hat{d}}=1$. An arithmetic mean in this context would yield $\mdhat \geq 0.5$, overestimating the similarity. In contrast, the harmonic mean results in $\mdhat=\epsilon$, where ($\epsilon \ll 1$) for most cases.

Note that while the PDDL problems $p_{1:N}$ and $\hat{p}_{1:N}$ appear in the definition of EW metrics, we only use the fact there are aligned object sets in them. We could also use an arbitrarily sampled object list to form an $\tilde{P}$ and pair $\tilde{P}$ with $D$ and $\hat{D}$ for EW metrics. But since for PDDL generation, we already generate $\hat{p}_{1:N}$, it is more convenient to use them.

Importantly, EW metrics can be computed without direct access to the full ground truth domain $d$ and problems $p$'s. As established before, to sample uniform random EW, we just need access to the object list and action interface, plus the environment executability checker of the source domain. So even for $m_{d \shortrightarrow \hat{d}}$, where the EW action sequences come from $d$, we do not need more than what is available through Assumption \ref{asm:env}.

To demonstrate the relationship between $\mdhat$ and domain disparity, we use the same simulated random omission study setup from Sec.\ \ref{sec:d_challenges}. For a pair of modified domains, we count the number of terms that differ, and inspect $\mdhat$ as function of increasing number of differing terms in Figure \ref{fig:ew_corr} for six example domains (see Figure \ref{fig:ew_corr_full} in the Appendix for the full set). We observe that, on average, a greater discrepancy in the number of terms between two domains correlates with a reduced EW score $\mdhat$. This observation provides additional support to the use of the EW score as an effective measure for domain differences.

\subsection{Leveraging LLMs to generate PDDL files}
We now show our overall LLM-based method for PDDL generation using the EW score to guide and measure the progress of domain generation. 
To illustrate the process, we first focus on a domain $d$ with a single task $p$. Recall that we are given NL description of the environment domain $d_\nl$ and problem $p_\nl$ (Assumption \ref{asm:description}), as well as the object list in $p$ and action interface from $d$ (Assumption \ref{asm:env}). Then, by using $d_\nl$, $p_\nl$, and access to environment action feedback, we seek to generate $\hat{d} \in \Dcal, \hat{p} \in \Pcal$. 

Our method starts by initializing templated $\hat{d}^{(0)}$ based on action interfaces and templated $\hat{p}^{(0)}$ using object list. Example template $\hat{d}^{(0)}$ and $\hat{p}^{(0)}$ are shown in Listings \ref{lst:grippers_domain_pddl_template} and \ref{lst:grippers_problem_template_pddl} of Appendix \ref{apx:dataset_details}. We then use an LLM to improve the initial $\hat{d}^{(0)}$ and $\hat{p}^{(0)}$. 

Given that domain PDDL files are typically more complex than problem PDDL files, our strategy prioritizes the generation of a problem PDDL file $\hat{p}$ first, followed by the domain $\hat{d}$. This approach enables us to assess the quality of the generated domain immediately. Moreover, prior works on code generation~\cite{codex}, tree-of-thought~\cite{yao2023tree}, and self-debug~\cite{self-debug} have found that taking multiple samples from the LLM response and taking the best response leads to better performance. However, they often require an evaluation metric on the generated response (such as unit test cases, or execution traces). Here, we use the EW metric introduced in Section \ref{sec:ew_intro} to serve as an evaluator of the generated domain. These considerations lead to our proposed Algorithm \ref{alg:method}. We emphasize again that the ground-truth domain and problem $d, p$ are only used to take exploration walks and evaluate a plan through the environment in \ref{alg:method}.

\begin{figure}
    \centering
    \includegraphics[width=.9\textwidth]{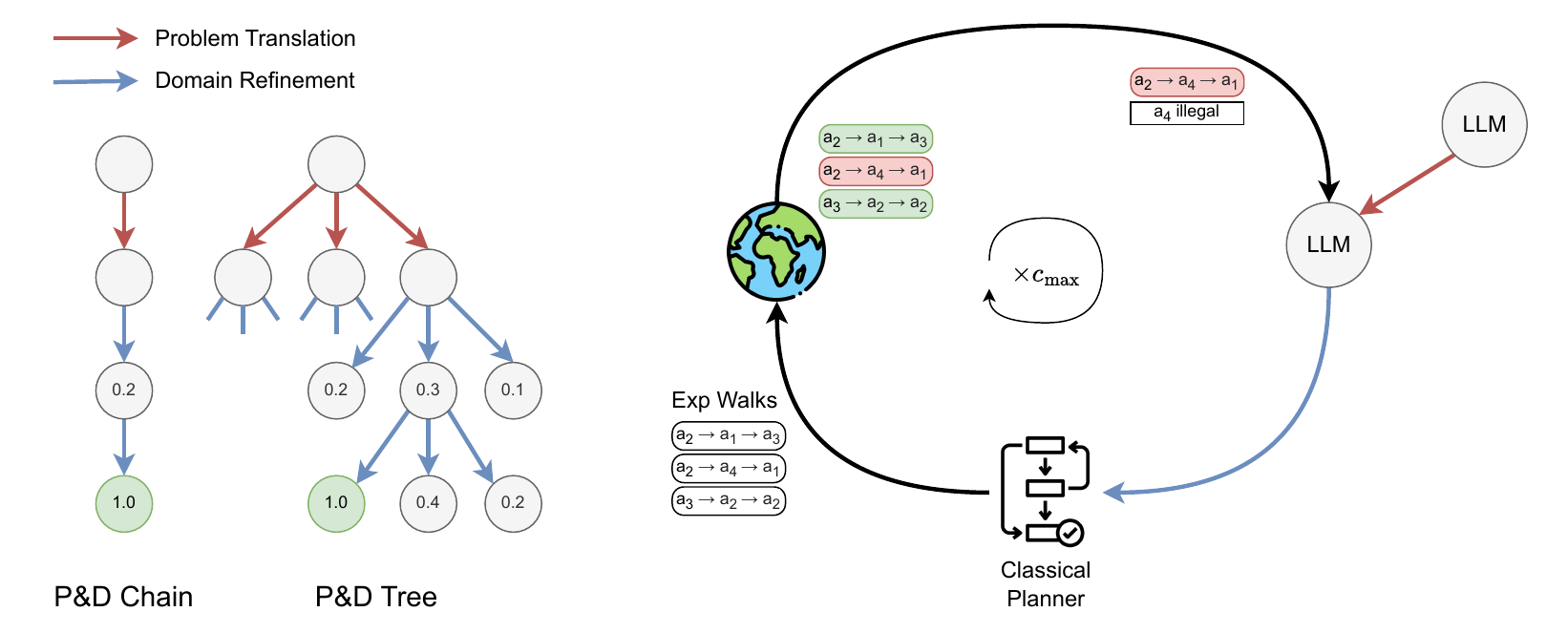}
    \caption{Overview of our method. \textit{Right:} The process begins with natural language descriptions translated into problem PDDL by the LLM (red arrows). Then a domain is generated and refined through iterative cycles involving exploration walks in the environment, interaction with a classical planner, and feedback from the LLM (blue/black arrows). \textit{Left:} The iterative refinement process depicted on the right corresponds to single paths in the structures shown on the left. Each node represents a state in the refinement process, with arrows indicating problem translation (red), domain refinement (blue).}
    \label{fig:method_comparison}
\end{figure}

\begin{algorithm}[h]
\caption{Generating Domain PDDL and Problem PDDL Using Environment Feedback}
\label{alg:method}
\begin{algorithmic}[1]
\REQUIRE Natural language descriptions $d_\nl$, $p_\nl$, environment action interface.
\STATE $\hat{p}^{(1)}, \hat{p}^{(2)}, \dots, \hat{p}^{(n_p)} \gets \llm_{n_p}(p_\nl)$ \COMMENT{Problem PDDL candidates}
\FOR{$i = 1, 2, \ldots, n_p$}
\STATE $h^{(i)} \gets [\hat{p}^{(i)}, d_\nl]$ \COMMENT{Keep a history of conversation}.
\STATE $\hat{d}_\text{best}^{(i)} \gets d_\nl$ \COMMENT{Initialize with an empty template}.
\FOR{$c = 1, 2, \ldots, c_\text{max}$}
\STATE $\hat{d}^{(i,1)}, \hat{d}^{(i,2)}, \dots, \hat{d}^{(i, n_d)} \gets \llm_{n_d}(h^{(i)})$
\STATE $\hat{d}^{(c)} \gets \argmax_{\hat{d} \in \{\hat{d}^{(i, 1)}, \dots, \hat{d}^{(i, n_d)} \}} \mdhat\left(p, \hat{p}^{(i)}\right)$ \COMMENT{Evaluate LLM responses using EW}
\STATE $f^{(c)} \gets$ Natural language feedback from EW on $d, p$.
\STATE $h^{(i)} \gets h^{(i)} + [\hat{d}^{(c)}, f^{(c)}]$
\STATE $\hat{d}_\text{best}^{(i)} \gets \argmax_{\hat{d} \in \{\hat{d}^{(c)}, \hat{d}_\text{best} \}} \mdhat\left(p, \hat{p}^{(i)}\right)$
\ENDFOR
\ENDFOR
\STATE $\hat{d}, \hat{p} \gets \argmax_{\{(\hat{d}_\text{best}^{(i)}, \hat{p}^{(i)}) \,|\, i = 1, 2, \dots, n_p\}} m_{d \leftrightarrow \hat{d}_\text{best}^{(i)}}(p, \hat{p}^{(i)})$
\RETURN $\hat{d}, \hat{p}$ \COMMENT{Return the final refined domain and problem PDDLs}
\end{algorithmic}
\end{algorithm}

Note that each environment contains $N>1$ problems, therefore, we need to translate all problem instances into PDDL. Similar to \citet{LLMP}, given one problem $p_{1_\text{NL}}$ and its generated translation $\hat{p}_1$, we translate the rest of the problems $p_{{2:N}_\text{NL}}$ in a one-shot manner. That is, we generate $\hat{p}_i := \llm_1 \big( p_{1_\text{NL}}, \hat{p}_1, p_{i_\text{NL}} \big)$ as the final problem translation for problem $i$ for all $2 \leq i \leq N$.

%% file: secs/4_experiments.tex
\section{Experiments}

\noindent\textbf{Dataset.} We consider PDDL files from real environments, taking nine domains from a combination of domain PDDLs from \citet{LLMP} and \citet{pddl_github}. 
The LLM may have seen the mentioned domains in its pre-training data, which is a common issue for current benchmarks. To mitigate this issue, we also modify the original Grippers domain, and create a modified domain called ``Grippers-ood'' domain, to ensure no LLM has seen it previously. We generate natural domain descriptions for all PDDL files by back-translating them using GPT-4 and manually inspecting and modifying the translations for correctness. For each environment, we consider one domain PDDL $d$ and $N=10$ problem PDDLs $p_{1:N}$. We use one problem for domain translation and EW evaluation, and all problems for evaluating a final domain response. We reserve the Blocksworld environment as an in-context example for prompting the $\llm$. As such, we do not evaluate the Blocksworld environment itself in our evaluations. See Appendices \ref{apx:dataset_details} and \ref{apx:translation_details} for more details on dataset curation.

\new{
\noindent\textbf{Feedback Format.}
The natural language feedback given to LLM is in the following form: \texttt{[Action sequence] [State description]}. That is, we first provide LLM with the sequence of actions taken from one exploration walk, up until one action fails. Then, we provide the environment state description from the last step.
We show an example of environment feedback and LLM response for the Termes environment in Listings \ref{lst:termes_feedback} in the Appendix. We deliberately choose a simple feedback format to maintain the general applicability of our framework.
}

\noindent\textbf{Baselines and Metrics.} We use GPT-4~\cite{gpt4} (\texttt{gpt-4-1106-preview}) as the $\llm$ since models with lower capability may struggle with syntax errors \cite{guan2023leveraging}. We consider the following methods: (1, 2) \textbf{Intrinsic Planning (CoT):} where the language model generates a complete plan without the help of any external planning library, based on the given descriptions, both with and without chain-of-thought prompting. This baseline does not leverage any classical planner or PDDL translation. (3) \textbf{P\&D Chain:} Our proposed method (Algorithm \ref{alg:method}) with $n_d = n_p = 1$. (4) \textbf{P\&D Tree:} Our proposed method with multiple response generations ($n_d = 10, n_p=5$). (5) \textbf{P\&D Tree + DomProp:} Our proposed method with multiple response generations and domain proposals for each problem (see Appendix \ref{apx:dom_prop}). \new{Following prior works~\cite{self_refine,self-debug}, we set a maximum conversation turns of $c_\text{max}=4$.}

We run each algorithm for four seeds and compute the Best@4 metric, which takes the highest score among the four seeds. We report two metrics: (1) tasks solved\footnote{\new{Note that a perfect task solve rate does not guarantee exact domain equivalency of the generated domain to the ground truth domain.}}, measuring the fraction of the $N=10$ tasks successfully solved (Eq. \eqref{eq:tasks_solved}), and (2) EW score, comparing the final domain through running exploration walks on all $N$ problems (Eq. \eqref{eq:ew_metric} with $T_\text{max} = 10$). We use the original fast-downward~\cite{fast_downward} library for planning, the modified fast-downward library from text-world~\cite{cote18textworld} for python-compatible state explorations, and the VAL~\cite{val_library} library to validate plans.

\noindent\textbf{Results.} Table \ref{tab:final_table} shows the final results on various environments. We consider a domain generation to be solved if a method achieves $>0.5$ solve rate since we observe the rest of the errors are problem translation errors rather than domain translation errors. Our proposed method solves 7 out of 10 domains, compared to 3 solved by the Intrinsic CoT baseline. We also generally observe the correlation of EW score with task solve rate. Particularly, even when the task solve rate is zero, the EW metric shows signs of progress, \eg, in domains such as Barman and Childsnack where all task solve rates are zero, the EW metric shows a clear distinction between method performances. Moreover, when the EW metric is high, such as $1.0$, we observe a generated PDDL domain to be very close to the ground-truth domain, and differing in very few predicates. For instance, in the case of the ``Hiking'' environment, the P\&D Chain achieves zero solve rate, but a perfect EW score, which we observe perfect solution in the case of P\&D Tree. 

\noindent\textbf{Computational Cost.} For the results in Table \ref{tab:final_table} using the GPT-4 model, we used $12.40$ million input tokens and $8.73$ million output tokens. \new{Computing the EW is relatively negligible compared to the cost of LLM inference. In our experiments, computing the EW score for a single domain-problem pair takes less than two minutes on a 64-core server CPU.}

\begin{table}
\small
\caption{Best@4 (Tasks solved / Exploration Walk) for different domains. For intrinsic planning no domain is generated, therefore the EW score is not defined.}
\label{tab:final_table}
\resizebox{\textwidth}{!}{%
\centering \begin{tabular}{lccccc}
\toprule
 & Intrinsic & Intrinsic & P\&D Chain & P\&D Tree & P\&D Tree + DomProp \\
 & No CoT & CoT & ($n_d=1, n_p=1$) & ($n_d=10, n_p=5$) & ($n_d=10, n_p=5$) \\
\midrule
Barman & 0.00 / -- & 0.00 / -- & 0.00 / 0.93 & 0.00 / 1.00 & 0.00 / 1.00 \\
Childsnack & 0.00 / -- & 0.00 / -- & 0.00 / 0.57 & 0.00 / 1.00 & 0.00 / 1.00 \\
Driverlog & 0.00 / -- & 0.00 / -- & 0.00 / 0.05 & 0.00 / 0.05 & 0.00 / 0.60 \\
Floortile & 0.00 / -- & 0.00 / -- & 0.00 / 0.07 & 0.90 / 0.94 & 0.00 / 0.07 \\
Grippers & 0.40 / -- & 0.60 / -- & 0.10 / 0.39 & 1.00 / 1.00 & 1.00 / 1.00 \\
Grippers-ood & 0.30 / -- & 0.30 / -- & 0.30 / 0.35 & 0.70 / 0.72 & 1.00 / 1.00 \\
Hiking & 0.00 / -- & 0.00 / -- & 0.00 / 1.00 & 1.00 / 1.00 & 1.00 / 1.00 \\
Miconic & 0.90 / -- & 1.00 / -- & 1.00 / 0.84 & 1.00 / 0.85 & 1.00 / 1.00 \\
Movie & 1.00 / -- & 1.00 / -- & 1.00 / 0.07 & 1.00 / 0.85 & 1.00 / 0.86 \\
Termes & 0.00 / -- & 0.00 / -- & 1.00 / 1.00 & 1.00 / 1.00 & 1.00 / 1.00 \\
\midrule \textbf{Average} & 0.26 / -- & 0.29 / -- & 0.34 / 0.53 & \textbf{0.66} / 0.84 & 0.60 / \textbf{0.85} \\
\bottomrule
\end{tabular}
}
\end{table}

%% file: secs/5_conclusion.tex
\section{Conclusion}
In this work, we present a novel approach for modeling planning environments via PDDL translation using large language models (LLMs) and environment feedback, without relying on human intervention. The key contributions include introducing the Exploration Walk (EW) metric to measure domain similarity and guide domain refinement, and an iterative method that leverages LLMs to generate and refine PDDL domain and problem files. Evaluation on 10 real-world PDDL domains demonstrates the effectiveness of the proposed approach, outperforming a baseline that generates PDDL files in a single attempt without refinement. The method solves 7 out of 10 environments, achieving an average task solve rate of 66\% and an average EW score of 0.84.

The current limitations include potentially insufficient and efficient exploration caused by random EW. More sophisticated EW strategies could improve the success rate while lowering the cost in the future. \new{For example, strategies from the reinforcement learning literature (\eg, \citep{rl-book, sagar2024failures}) could be adapted to improve exploration efficiency and success rates.} Another limitation is that we have only applied the framework to PDDL environments, despite it being applicable to digital or even physical environments. We hope this work will inspire further research at the intersection of language models and planning, enabling the development of more advanced and autonomous planning systems.

%% file: secs/appendix.tex
\begin{table}[h!]
    \caption{Summary of Notation and Definitions}
    \label{tab:notation}
    \centering
    \begin{tabular}{lp{9cm}}
        \toprule
        \textbf{Notation} & \textbf{Description} \\
        \midrule
        $1:N$ & Sequence of integers ranging from $1$ to $N$ \\
        \midrule
        $\Acal^*$ & Set comprising all possible sequences of elements drawn from set $\Acal$ \\
        \midrule
        $2^\Acal$ & Power set of $\Acal$ \\
        \midrule
        $\Dcal$ & Set of all possible domains in PDDL \\
        \midrule
        $\Pcal$ & Set of all possible problems in PDDL \\
        \midrule
        $\Acal$ & Set of all possible actions in PDDL \\
        \midrule
        $\planerr$ & Planning error \\
        \midrule
        $C: \Dcal \times \Pcal \rightarrow \Acal^*\cup \{\planerr\}$ & Classical planner function that takes a domain $d \in \Dcal$ and a problem $p \in \Pcal$ and produces a plan $q$ \\
        \midrule
        $V_{d, p}(q): \Acal^* \rightarrow \{0, 1\}$ & Plan validator function for domain $d$ and problem $p$, returns $1$ if plan $q$ is valid, otherwise $0$ \\
        \midrule
        $E_{d, p}: \Acal^* \rightarrow \{0, 1\}$ & Plan execution checker for domain $d$ and problem $p$, returns $1$ if action sequence is executable, otherwise $0$ \\
        \midrule
        $\Scal$ & Set of all possible states \\
        \midrule
        $A_{d, p}: \Scal \rightarrow 2^\Acal$ & Function delineating the set of legal actions given the current state for domain $d$ and problem $p$ \\
        \midrule
        $S_{d, p}: \Acal \times \Scal \rightarrow \Scal$ & State transition function, determines the next state given the current state and action in domain $d$ and problem $p$ \\
        \midrule
        $s_{d, p, 0}$ & Initial state induced by domain $d$ and problem $p$ \\
        \midrule
        $\llm_n(X)$ & Sampling $n$ responses from the $\llm$ given prompt $X$ \\
        \bottomrule
    \end{tabular}
\end{table}

\section{Dataset}

\begin{figure}[t]
    \centering
    \includegraphics[width=\textwidth]{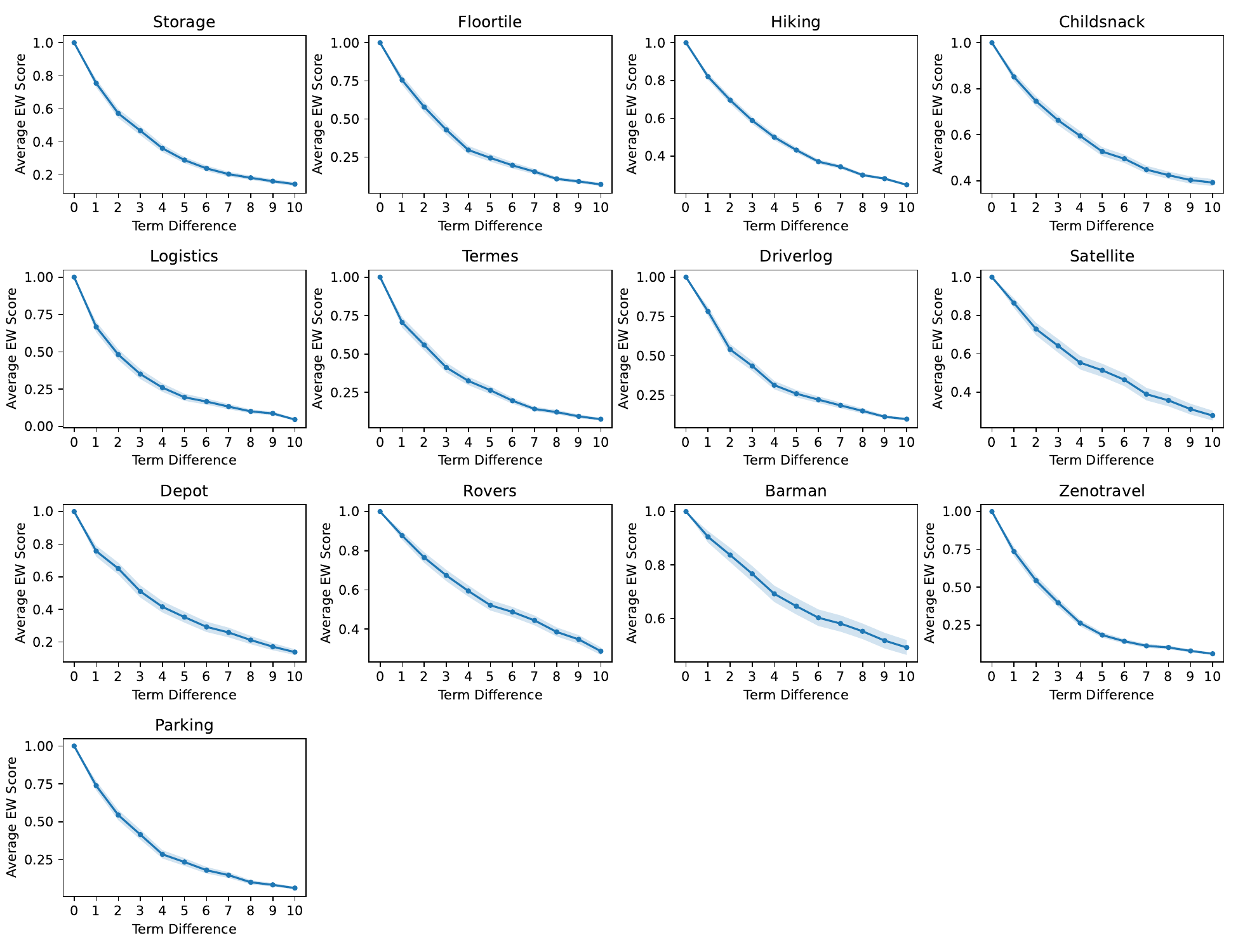}
    \caption{Correlation between average exploration walk score and average domain difference}
    \label{fig:ew_corr_full}
\end{figure}

\subsection{Dataset Details.} \label{apx:dataset_details}

\textbf{Dataset Examples.} We provide an example of each file for the Grippers environment: (1) The ground-truth domain  $d$ (Listing \ref{lst:grippers_domain_pddl}) of ground truth PDDL domain (2) One ground-truth problem $p$ (Listing \ref{lst:grippers_problem_pddl}) (3) Domain natural language description along with a PDDL template for action interfaces $d_\text{NL}$ (Listings \ref{lst:grippers_domain_nl} and \ref{lst:grippers_domain_pddl_template}) (4) Problem natural language description along with a PDDL template with the list of objects (Listings \ref{lst:grippers_p01_nl} and \ref{lst:grippers_problem_template_pddl})

\input{figs/grippers_pddl}
\input{figs/grippers_p01_pddl}
\input{figs/grippers_p01_nl}
\input{figs/grippers_p01_pddl_template}
\input{figs/grippers_nl}
\input{figs/grippers_pddl_template}
\input{figs/grippers_p01_plan}

\begin{figure}[t]
    \centering
    \includegraphics[width=0.5\textwidth]{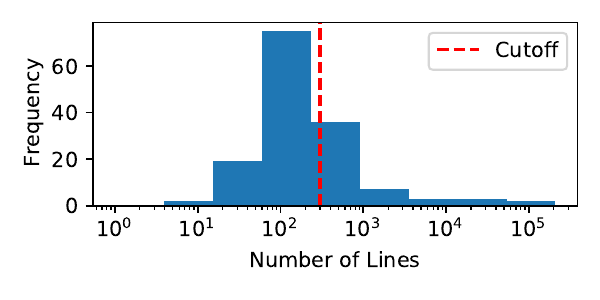}
    \caption{Historgram of average number of lines of domains in \cite{pddl_github}.}
    \label{fig:domain_hist}
\end{figure}

\subsection{Criticality of predicate design.} \label{apx:predicate_design}
Here, we give an example on the delicacy of predicate design. Consider the Grippers environment, where each robot has two grippers: left gripper and right gripper. In our experiments, one of the main predicates that the LLM incorrectly generates is the \texttt{free} predicate (see Listing \ref{lst:grippers_problem_pddl_wrong}). This predicate keeps track of whether a gripper is free or not. Therefore, at first sight, \texttt{(free ?g - gripper)} seems a natural choice to show a particular gripper is not occupied and hence is capable of picking a ball. However, when designed this way, in contrast to \texttt{(free ?r - robot ?g - gripper)} (missing the robot argument), this small detail causes the final domain to be entirely wrong! The reason is that there would no longer be any association between a robot and its two grippers. Therefore, on the incorrect domain, one robot will be able to pickup an object with the gripper of another robot! In fact, we observe that this incorrect design for the \texttt{free} predicate, is the reason behind the failure of the ``P\&D Chain'' method in Table \ref{tab:final_table}.

\new{We provide one more example from the Barman environment, illustrating the criticality of predicate design. The Barman environment involves actions related to manipulating containers (e.g., shot glasses, shakers) to prepare and serve drinks using various ingredients. One of the key predicates used in the domain is \texttt{(used ?c - container ?b - beverage)}, which keeps track of which beverage has been used in a specific container. This is important for actions like refilling or cleaning, where knowing the specific beverage type is essential to ensure conformation to the environment rules (\eg, a container can be refilled only with the beverage that it already had, otherwise, it needs to be cleaned first).
However, we have observed that when the LLM generates the domain, it sometimes mistakenly omits the beverage argument, simplifying the predicate to \texttt{(used ?c - container)}. At first glance, this might seem like a harmless simplification, as the container usage is still tracked. However, this change results in significant problems in the overall domain behavior. Since the beverage is no longer specified, the domain can no longer differentiate between containers used for different types of beverages. This leads to situations where a container that has already been used for one beverage can be incorrectly treated as if it can hold another beverage without requiring proper cleaning or resetting actions.
Such a mistake can cause the final domain to generate invalid plans, as the planner will fail to ensure that containers are used properly with respect to their contents, leading to cascading errors in tasks like mixing drinks, cleaning containers, or pouring from shakers.}

\input{figs/grippers_p01_pddl_wrong}

\section{Implementation Details} \label{apx:method_details}
In this section, we explain our implementation details.

\subsection{One-shot prompting} To generate PDDL files (problem PDDL and domain PDDL), we always include a one-shot example prompt from the BlocksWorld environment. 
This environment is concise easy enough to fit into context, and explanatory enough to demonstrate example to the LLM for better output steerability.
This includes problem generation, domain proposal, and problem refinement.
For instance, when prompting the $\llm$ to generate problem translation from natural language, \eg, $\llm(p_\text{NL})$, we also prompt the $\llm$ with an example from Blocksworld.

\subsection{P\&D Tree with Domain Proposal} \label{apx:dom_prop}
As discussed in \ref{apx:predicate_design}, predicate design is challenging. Therefore, in one variant of our method, which we call ``P\&D Tree DomProp'', we propose for the LLM to first draft a domain proposal, then generate a problem PDDL based on the predicates found in the draft. This way, the LLM first generates domain-aware predicates, then generates the problem PDDL. Formally, line one in Algorithm \ref{alg:method} will be changed to the following two lines:
\begin{algorithmic}
\STATE $\hat{d}_\text{pr}^{(1)}, \hat{d}_\text{pr}^{(2)}, \dots, \hat{d}_\text{pr}^{(n_p)} \gets \llm_{n_p}(d_\nl)$
\STATE $\hat{p}^{(i)} \gets \llm_{1}(\hat{d}_\text{pr}^{(i)}, p_\nl) ~~\algorithmicforall~~1 \leq i \leq n_p$
\end{algorithmic}
where the problem PDDL is generated by first creating a domain proposal.

\begin{figure}
    \centering
    \includegraphics[width=.6\textwidth]{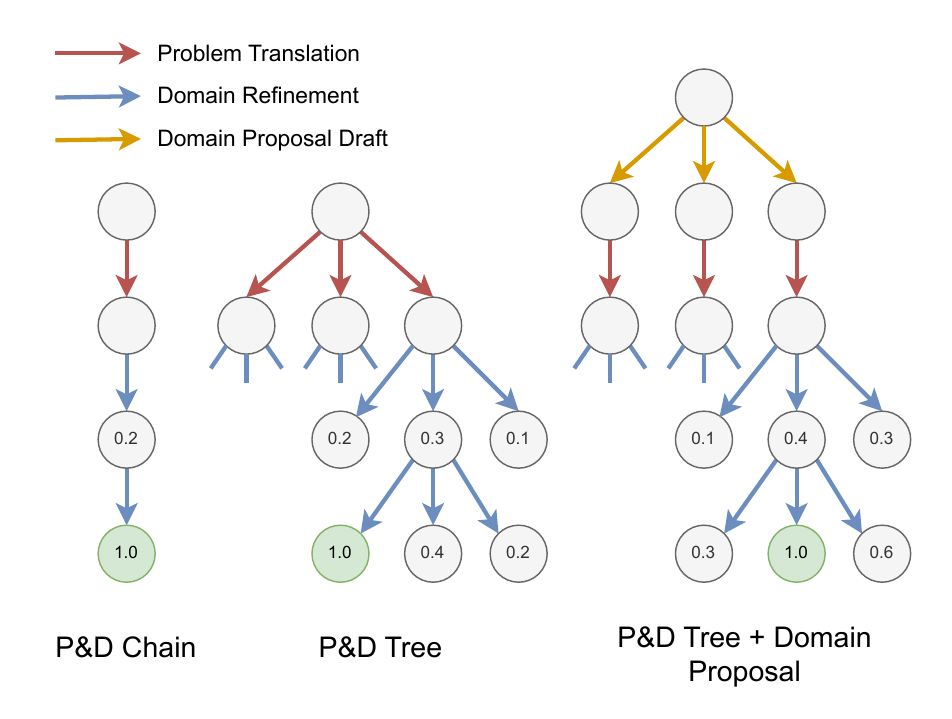}
    \caption{Overview of our method with domain proposal. To generate a problem PDDL, the LLM first drafts a domain proposal to find suitable predicates for the problem PDDL. Then, the draft is discarded, and the domain refinement stage starts.}
    \label{fig:method_comparison}
\end{figure}

\subsection{Domain Refinement Strategy} 
\textbf{Refinement Interface.} For the domain refinement stage, in our early experiments we observed that prompting the LLM to regenerate the domain results into redundant output generation and more importantly, sometimes modifies incorrect parts of the domain.
For instance, the LLM had a high tendency towards changing the action interface signature, despite the instructions explicitly mentioning not to change the signature.
As such, we provide a python interface for the LLM to modify a domain. The interface provides the LLM with the following two functions:

\begin{lstlisting}[language=Python]
add_or_update_predicates(
    predicates: List[str]
)
modify_action(
    action_name: str, 
    new_preconditions: List[str], 
    new_effects: List[str]
)
\end{lstlisting}
The first function adds predicates to the list of already created predicates, and the second one modifies the preconditions and effects of a particular action. \citet{guan2023leveraging} use a similar approach where they generate the domain PDDL one action at a time, and gradually create predicates. 
However, our python function interface allows for more flexibility, such as more convenient implementation as well as enabling the LLM to modify an action several times, or introduce predicates in between reasoning steps.

\textbf{Domain Rating.} Our main domain rating originates from the EW metric. When generating domain refinement strategies, the LLM may make mistakes hence failing before even the EW metric could be computed. For instance, the modification may be invalid, containing syntax error, or failing to fill parts of the template. To facilitate incorporating these into the EW metric strategy, we create the following rating system for each domain refinement modification:

\begin{center}
\begin{tabular}{lc}
\toprule
\textbf{Description} & \textbf{Rating} \\ 
\midrule
Exploration Walk Executable & $0 \leq \text{EW Score} \leq 1$ \\ 
\midrule
No initial action possible & $-1$ \\ 
\midrule
Invalid domain modification (\eg, undefined predicates) & $-2$ \\ 
\midrule
Domain sanity check failure (\eg, empty effect list) & $-3$ \\ 
\midrule
Invalid domain modification & $-4$ \\ 
\midrule
No domain modification & $-5$ \\ 
\bottomrule
\end{tabular}
\end{center}

when the EW metric is perfect (\ie, equals $1.0$), we also run the planning on the environment (\ie, evaluate $V_{d, p}(.)$) and stop early if the plan is valid.

\input{figs/termes_feedback}

\subsection{Further experiment details}
\new{
\textbf{LLM calls per task.} For each task in Algorithm \ref{alg:method}, the overall complexity of LLM calls is $O(n_p \times n_d \times c_{max} + N )$. The complexity of domain generation is $O(n_p \times c_{\text{max}} \times n_d)$. This is because at first, $n_p$ problem candidates are generated and for each problem candidate the algorithm goes through a refinement procedure (lines 1 and 2 of Algorithm \ref{alg:method}). The refinement is a tree with depth $c_{\text{max}}$ (where $c_{\text{max}}$ is the maximum number of refinement turns) (line 5), and at each level of the tree, one node is expanded with $n_d$ children (where $n_d$ is the number of domain refinement candidates) (line 6), which leads to $O(n_p \times c_{\text{max}} \times n_d)$ complexity. Once the domain is ready, the complexity of task generation for $N$ tasks is $O(N)$ since for each task we only call the LLM once to get a problem translation.
}

\new{
\textbf{Number of successful seeds.} In Table \ref{tab:final_table}, we report the results over four seeds. To provide further analysis, we report the number of seeds a domain was successful in successfully generating a correct domain. The number of seeds that succeed in generating correct domain for the Termes, Movie, Miconic, Grippers, Hiking, Grippers-ood, and Floortile, are 4, 3, 3, 3, 2, 1, 1, respectively. 
}

\section{Natural Language Description Generation} \label{apx:translation_details}

To generate natural language description of domains, problems, and environment states, we use the following strategies:
\begin{itemize}
    \item \textbf{Domain:} We use a few-shot translation strategy. We first pick three diverse environments of ``Grippers'', ``Childsnack'', and ``Termes'' to manually (with assistance of GPT-4) curate domain translation. Then, we use these three domains as three-shot in-context examples to translate the rest of domains. The example prompt is provided in Listing \ref{lst:prompt_domain_translation}.
    \item \textbf{Problems:} We use a similar few-shot translation strategy for problem translation. We first pick two diverse environments of ``Termes'' and ``Satellite'' for problem two-shot problem translation. Once one problem from a target domain is translated, we use that problem translation as in-context example to translate the rest of the problems. This step is crucial to ensure all problems from the same domain are translated in a consistent manner. The example prompt is provided in Listing \ref{lst:prompt_problem_translation}.
    \item \textbf{Natural Language Predicate Description:} To generate natural language description of states, we generate a python files for each domain, with one function to produce natural language description of predicates for state description. The example prompt is provided in Listing \ref{lst:prompt_predicate_translation}.
\end{itemize}

\input{figs/prompt_domain_translation}
\input{figs/prompt_problem_translation}
\input{figs/prompt_predicate_desc}

%% file: figs/grippers_pddl.tex
\begin{lstlisting}[language=pddl, caption={Grippers domain PDDL~\cite{LLMP}.}, label={lst:grippers_domain_pddl}]
(define (domain gripper-strips)
  (:requirements :strips :typing)
  (:types room obj robot gripper)
  (:predicates (at-robby ?r - robot ?x - room)
               (at ?o - obj ?x - room)
               (free ?r - robot ?g - gripper)
               (carry ?r - robot ?o - obj ?g - gripper))

  (:action move
    :parameters (?r - robot ?from ?to - room)
    :precondition (and (at-robby ?r ?from))
    :effect (and (at-robby ?r ?to)
                 (not (at-robby ?r ?from))))

  (:action pick
    :parameters (?r - robot ?obj - obj ?room - room ?g - gripper)
    :precondition (and (at ?obj ?room) (at-robby ?r ?room) (free ?r ?g))
    :effect (and (carry ?r ?obj ?g)
                 (not (at ?obj ?room))
                 (not (free ?r ?g))))

  (:action drop
    :parameters (?r - robot ?obj - obj ?room - room ?g - gripper)
    :precondition (and (carry ?r ?obj ?g) (at-robby ?r ?room))
    :effect (and (at ?obj ?room)
                 (free ?r ?g)
                 (not (carry ?r ?obj ?g)))))
\end{lstlisting}

%% file: figs/grippers_p01_pddl.tex
\begin{lstlisting}[language=pddl, caption={Grippers problem PDDL.}, label={lst:grippers_problem_pddl}]
(define (problem gripper-2-3-4)
  (:domain gripper-strips)
  (:objects robot1 robot2 - robot
            rgripper1 lgripper1 rgripper2 lgripper2 - gripper
            room1 room2 room3 - room
            ball1 ball2 ball3 ball4 - obj)
  (:init
    (at-robby robot1 room2)
    (free robot1 rgripper1)
    (free robot1 lgripper1)
    (at-robby robot2 room3)
    (free robot2 rgripper2)
    (free robot2 lgripper2)
    (at ball1 room3)
    (at ball2 room1)
    (at ball3 room1)
    (at ball4 room3)
  )
  (:goal
    (and
      (at ball1 room2)
      (at ball2 room2)
      (at ball3 room3)
      (at ball4 room3)
    )
  )
)
\end{lstlisting}

%% file: figs/grippers_p01_nl.tex
\begin{lstlisting}[language=markdown, caption={Grippers problem natural language.}, label={lst:grippers_p01_nl}]
You control two robots, each equipped with a left and right gripper, capable of moving objects (balls) between different rooms.

Initially:
- Robot1 is in room2 and both its grippers (rgripper1 and lgripper1) are free.
- Robot2 is in room3 and both its grippers (rgripper2 and lgripper2) are free.
- Ball1 and Ball4 are in room3.
- Ball2 and Ball3 are in room1.

Your goal is to achieve the following configuration:
- Ball1 must be moved to room2.
- Ball2 must be moved to room2.
- Ball3 must remain in room3.
- Ball4 must remain in room3.
\end{lstlisting}

%% file: figs/grippers_p01_pddl_template.tex
\begin{lstlisting}[language=pddl, caption={Grippers problem template PDDL.}, label={lst:grippers_problem_template_pddl}]
(define (problem gripper-2-3-4)
    (:domain gripper-strips)
    (:objects lgripper1 lgripper2 rgripper1 rgripper2 - gripper ball1 ball2 ball3 ball4 - obj robot1 robot2 - robot room1 room2 room3 - room)
    (:init )
    (:goal (and ))
)
\end{lstlisting}

%% file: figs/grippers_nl.tex
\begin{lstlisting}[language=markdown, caption={Grippers domain natural language.}, label={lst:grippers_domain_nl}]
The gripper domain involves a world with multiple rooms, robots, and objects (balls). Each robot has two grippers that can be used to pick up and drop objects. The goal is to move objects from their initial locations to the desired goal locations using the robots and their grippers.

The domain includes three actions:

1. move: This action allows a robot to move from one room to another. The precondition is that the robot must be in the starting room. The effect is that the robot is no longer in the starting room and is now in the destination room.

2. pick: This action allows a robot to pick up an object using one of its grippers. The preconditions are that the object and the robot must be in the same room, and the specified gripper must be free (not holding any object). The effect is that the robot is now carrying the object with the specified gripper, the object is no longer in the room, and the gripper is no longer free.

3. drop: This action allows a robot to drop an object it is carrying in a specific room using one of its grippers. The preconditions are that the robot must be carrying the object with the specified gripper and the robot must be in the specified room. The effect is that the object is now in the room, the gripper is free, and the robot is no longer carrying the object with that gripper.
\end{lstlisting}

%% file: figs/grippers_pddl_template.tex
\begin{lstlisting}[language=pddl, caption={Grippers domain PDDL template.}, label={lst:grippers_domain_pddl_template}]
(define (domain gripper-strips)
  (:requirements :strips :typing)
  (:types room obj robot gripper)
  (:predicates)

  (:action move
    :parameters (?r - robot ?from ?to - room)
    :precondition ()
    :effect ())

  (:action pick
    :parameters (?r - robot ?o - obj ?room - room ?g - gripper)
    :precondition ()
    :effect ())

  (:action drop
    :parameters (?r - robot ?o - obj ?room - room ?g - gripper)
    :precondition ()
    :effect ()))
\end{lstlisting}

%% file: figs/grippers_p01_plan.tex
\begin{lstlisting}[language=pddl, caption={Grippers problem plan example.}, label={lst:grippers_problem_plan}]
(move robot2 room3 room1)
(pick robot2 ball2 room1 lgripper2)
(move robot2 room1 room2)
(drop robot2 ball2 room2 lgripper2)
(move robot1 room2 room1)
(pick robot1 ball3 room1 lgripper1)
(move robot1 room1 room3)
(pick robot1 ball1 room3 rgripper1)
(drop robot1 ball3 room3 lgripper1)
(move robot1 room3 room2)
(drop robot1 ball1 room2 rgripper1)
\end{lstlisting}

%% file: figs/grippers_p01_pddl_wrong.tex
\begin{lstlisting}[language=pddl, caption={Incorrect generated grippers problem PDDL. The \texttt{free} predicate has only one parameter.}, label={lst:grippers_problem_pddl_wrong}]
(define (problem gripper-2-3-4)
  (:domain gripper-strips)
  (:objects robot1 robot2 - robot
            rgripper1 lgripper1 rgripper2 lgripper2 - gripper
            room1 room2 room3 - room
            ball1 ball2 ball3 ball4 - obj)
  (:init
    (at-robby robot1 room2)
    (free rgripper1) ; Correct: (free (*@\highlb{robot1}@*) rgripper1)
    (free lgripper1) ; Correct: (free (*@\highlb{robot1}@*) lgripper1)
    (at-robby robot2 room3)
    (free rgripper2) ; Correct: (free (*@\highlb{robot2}@*) rgripper2)
    (free lgripper2) ; Correct: (free (*@\highlb{robot2}@*) lgripper2)
    (at ball1 room3)
    (at ball2 room1)
    (at ball3 room1)
    (at ball4 room3)
  )
  (:goal
    (and
      (at ball1 room2)
      (at ball2 room2)
      (at ball3 room3)
      (at ball4 room3)
    )
  )
)
\end{lstlisting}

%% file: figs/termes_feedback.tex
\begin{lstlisting}[language=markdown, caption={\new{Example of domain feedback from the Termes environment, where the LLM output refinement results into a correct domain.}}, label={lst:termes_feedback}]
(*@\highly{User:}@*)
Incorrect. The environment returned the following error:

Sampled a set of consecutive random actions from the ground truth environment, but the actions are not executable in the generated environment.
Executing the following actions sequentially on the environment:
(move pos-2-0 pos-3-0 n0)
(move pos-3-0 pos-2-0 n0)
(create-block pos-2-0)
(move pos-2-0 pos-3-0 n0)
(move pos-3-0 pos-2-0 n0)
(place-block pos-2-0 pos-1-0 n0 n1)
(remove-block pos-2-0 pos-1-0 n1 n0)
(place-block pos-2-0 pos-2-1 n0 n1)
(remove-block pos-2-0 pos-2-1 n1 n0)
(place-block pos-2-0 pos-3-0 n0 n1)
(create-block pos-2-0)
(place-block pos-2-0 pos-2-1 n0 n1)
(create-block pos-2-0)
(place-block pos-2-0 pos-1-0 n0 n1)
(create-block pos-2-0)
(move-up pos-2-0 n0 pos-1-0 n1)

Result: Error when executing the action (move-up pos-2-0 n0 pos-1-0 n1).
Current state: The robot is at position pos-2-0.
This action is not executable on the environment.

Please reason about the issue with your generated code. The current domain pddl is as follows:

```pddl
(*@\highlb{[Current Generated PDDL]}@*)
```

In your response, please generate a new code to fix the issue.

(*@\highly{Assistant:}@*)
It seems that there is an issue with the `move-up` action precondition. The precondition for `move-up` should check that the destination position `?hto` is one less than the starting position `?hfrom`, not the other way around. This is because the robot is moving to a higher position, so the destination height should be greater than the starting height.

Similarly, the precondition for `move-down` should check that the destination position `?hto` is one less than the starting position `?hfrom`, indicating that the robot is moving to a lower position.

Let's correct the preconditions for `move-up` and `move-down`:

```python
(*@\highlb{[Python Domain Refinement Function Calls]}@*)
```

With these corrected preconditions, the `move-up` and `move-down` actions should now accurately reflect the natural language description of the Termes domain, and the error should be resolved.
\end{lstlisting}

%% file: figs/prompt_domain_translation.tex
\begin{lstlisting}[language=markdown, caption={Domain back-translation prompt template, with domain PDDL and problem PDDL placeholders for each target domain. Some PDDL credit comments are omitted for clarity. }, label={lst:prompt_domain_translation}]
Your task is to translate PDDL files into natural language. 
Ensure that the resulting text covers natural language description of its actions, their preconditions, and effects.
DO NOT translate the problem PDDL files, only use problem PDDL to understand the domain. ALWAYS wrap your code in the appropriate markdown syntax.
Two examples are provided below.
Q:
Domain PDDL:
```pddl
(define (domain gripper-strips)
 (:requirements :strips :typing) 
 (:types room obj robot gripper)
 (:predicates (at-robby ?r - robot ?x - room)
 	      (at ?o - obj ?x - room)
	      (free ?r - robot ?g - gripper)
	      (carry ?r - robot ?o - obj ?g - gripper))
   (:action move
       :parameters  (?r - robot ?from ?to - room)
       :precondition (and  (at-robby ?r ?from))
       :effect (and  (at-robby ?r ?to)
		     (not (at-robby ?r ?from))))
   (:action pick
       :parameters (?r - robot ?obj - obj ?room - room ?g - gripper)
       :precondition  (and  (at ?obj ?room) (at-robby ?r ?room) (free ?r ?g))
       :effect (and (carry ?r ?obj ?g)
		    (not (at ?obj ?room)) 
		    (not (free ?r ?g))))
   (:action drop
       :parameters (?r - robot ?obj - obj ?room - room ?g - gripper)
       :precondition  (and  (carry ?r ?obj ?g) (at-robby ?r ?room))
       :effect (and (at ?obj ?room)
		    (free ?r ?g)
		    (not (carry ?r ?obj ?g)))))
```
Problem PDDL:
```pddl
(define (problem gripper-2-4-6)
(:domain gripper-strips)
(:objects robot1 robot2 - robot
rgripper1 lgripper1 rgripper2 lgripper2 - gripper
room1 room2 room3 room4 - room
ball1 ball2 ball3 ball4 ball5 ball6 - obj)
(:init
(at-robby robot1 room2)
(free robot1 rgripper1)
(free robot1 lgripper1)
(at-robby robot2 room3)
(free robot2 rgripper2)
(free robot2 lgripper2)
(at ball1 room3)
(at ball2 room1)
(at ball3 room3)
(at ball4 room2)
(at ball5 room4)
(at ball6 room4)
)
(:goal
(and
(at ball1 room4)
(at ball2 room1)
(at ball3 room1)
(at ball4 room2)
(at ball5 room1)
(at ball6 room1)
)
)
)
```
A:
```markdown
The gripper domain involves a world with multiple rooms, robots, and objects (balls). Each robot has two grippers that can be used to pick up and drop objects. The goal is to move objects from their initial locations to the desired goal locations using the robots and their grippers.
The domain includes three actions:
1. move: This action allows a robot to move from one room to another. The precondition is that the robot must be in the starting room. The effect is that the robot is no longer in the starting room and is now in the destination room.
2. pick: This action allows a robot to pick up an object using one of its grippers. The preconditions are that the object and the robot must be in the same room, and the specified gripper must be free (not holding any object). The effect is that the robot is now carrying the object with the specified gripper, the object is no longer in the room, and the gripper is no longer free.
3. drop: This action allows a robot to drop an object it is carrying in a specific room using one of its grippers. The preconditions are that the robot must be carrying the object with the specified gripper and the robot must be in the specified room. The effect is that the object is now in the room, the gripper is free, and the robot is no longer carrying the object with that gripper.
```
Q:
Domain PDDL:
```pddl
(define (domain child-snack)
(:requirements :typing :equality)
(:types child bread-portion content-portion sandwich tray place)
(:constants kitchen - place)
(:predicates (at_kitchen_bread ?b - bread-portion)
	     (at_kitchen_content ?c - content-portion)
     	     (at_kitchen_sandwich ?s - sandwich)
     	     (no_gluten_bread ?b - bread-portion)
       	     (no_gluten_content ?c - content-portion)
      	     (ontray ?s - sandwich ?t - tray)
       	     (no_gluten_sandwich ?s - sandwich)
	     (allergic_gluten ?c - child)
     	     (not_allergic_gluten ?c - child)
	     (served ?c - child)
	     (waiting ?c - child ?p - place)
             (at ?t - tray ?p - place)
	     (notexist ?s - sandwich)
  )
(:action make_sandwich_no_gluten 
	 :parameters (?s - sandwich ?b - bread-portion ?c - content-portion)
	 :precondition (and (at_kitchen_bread ?b)
			    (at_kitchen_content ?c)
			    (no_gluten_bread ?b)
			    (no_gluten_content ?c)
			    (notexist ?s))
	 :effect (and
		   (not (at_kitchen_bread ?b))
		   (not (at_kitchen_content ?c))
		   (at_kitchen_sandwich ?s)
		   (no_gluten_sandwich ?s)
                   (not (notexist ?s))
		   ))
(:action make_sandwich
	 :parameters (?s - sandwich ?b - bread-portion ?c - content-portion)
	 :precondition (and (at_kitchen_bread ?b)
			    (at_kitchen_content ?c)
                            (notexist ?s)
			    )
	 :effect (and
		   (not (at_kitchen_bread ?b))
		   (not (at_kitchen_content ?c))
		   (at_kitchen_sandwich ?s)
                   (not (notexist ?s))
		   ))
(:action put_on_tray
	 :parameters (?s - sandwich ?t - tray)
	 :precondition (and  (at_kitchen_sandwich ?s)
			     (at ?t kitchen))
	 :effect (and
		   (not (at_kitchen_sandwich ?s))
		   (ontray ?s ?t)))
(:action serve_sandwich_no_gluten
 	:parameters (?s - sandwich ?c - child ?t - tray ?p - place)
	:precondition (and
		       (allergic_gluten ?c)
		       (ontray ?s ?t)
		       (waiting ?c ?p)
		       (no_gluten_sandwich ?s)
                       (at ?t ?p)
		       )
	:effect (and (not (ontray ?s ?t))
		     (served ?c)))
(:action serve_sandwich
	:parameters (?s - sandwich ?c - child ?t - tray ?p - place)
	:precondition (and (not_allergic_gluten ?c)
	                   (waiting ?c ?p)
			   (ontray ?s ?t)
			   (at ?t ?p))
	:effect (and (not (ontray ?s ?t))
		     (served ?c)))
(:action move_tray
	 :parameters (?t - tray ?p1 ?p2 - place)
	 :precondition (and (at ?t ?p1))
	 :effect (and (not (at ?t ?p1))
		      (at ?t ?p2)))
			    
)
```
Problem PDDL:
```pddl
; child-snack task with 6 children and 0.4 gluten factor 
; constant factor of 1.3
; random seed: 234324
(define (problem prob-snack)
  (:domain child-snack)
  (:objects
    child1 child2 child3 child4 child5 child6 - child
    bread1 bread2 bread3 bread4 bread5 bread6 - bread-portion
    content1 content2 content3 content4 content5 content6 - content-portion
    tray1 tray2 - tray
    table1 table2 table3 - place
    sandw1 sandw2 sandw3 sandw4 sandw5 sandw6 sandw7 sandw8 - sandwich
  )
  (:init
     (at tray1 kitchen)
     (at tray2 kitchen)
     (at_kitchen_bread bread1)
     (at_kitchen_bread bread2)
     (at_kitchen_bread bread3)
     (at_kitchen_bread bread4)
     (at_kitchen_bread bread5)
     (at_kitchen_bread bread6)
     (at_kitchen_content content1)
     (at_kitchen_content content2)
     (at_kitchen_content content3)
     (at_kitchen_content content4)
     (at_kitchen_content content5)
     (at_kitchen_content content6)
     (no_gluten_bread bread2)
     (no_gluten_bread bread5)
     (no_gluten_content content3)
     (no_gluten_content content6)
     (allergic_gluten child1)
     (allergic_gluten child5)
     (not_allergic_gluten child2)
     (not_allergic_gluten child3)
     (not_allergic_gluten child4)
     (not_allergic_gluten child6)
     (waiting child1 table2)
     (waiting child2 table1)
     (waiting child3 table3)
     (waiting child4 table2)
     (waiting child5 table1)
     (waiting child6 table1)
     (notexist sandw1)
     (notexist sandw2)
     (notexist sandw3)
     (notexist sandw4)
     (notexist sandw5)
     (notexist sandw6)
     (notexist sandw7)
     (notexist sandw8)
  )
  (:goal
    (and
     (served child1)
     (served child2)
     (served child3)
     (served child4)
     (served child5)
     (served child6)
    )
  )
)
```
A:
```markdown
The child-snack domain is designed to plan the preparation and serving of sandwiches to children, some of whom have gluten allergies. The domain includes actions for making sandwiches with and without gluten, placing sandwiches on trays, moving trays between places, and serving the sandwiches to the children.
The actions defined in this domain include:
1. make_sandwich_no_gluten: This action is for making a gluten-free sandwich. The preconditions for this action are that there must be gluten-free bread and gluten-free content available in the kitchen, and the sandwich must not already exist. The effect of this action is that the gluten-free bread and content are no longer available in the kitchen, and a gluten-free sandwich is now available in the kitchen.
2. make_sandwich: This action is for making a regular sandwich that may contain gluten. The preconditions are that there must be bread and content available in the kitchen, and the sandwich must not already exist. The effect of this action is that the bread and content are no longer available in the kitchen, and a sandwich is now available in the kitchen.
3. put_on_tray: This action is for placing a sandwich on a tray. The preconditions are that the sandwich must be in the kitchen and the tray must be at the kitchen. The effect is that the sandwich is no longer in the kitchen and is now on the tray.
4. serve_sandwich_no_gluten: This action is for serving a gluten-free sandwich to a child who is allergic to gluten. The preconditions are that the child is allergic to gluten, waiting at a place, and there is a gluten-free sandwich on a tray at the same place. The effect is that the sandwich is no longer on the tray and the child is served.
5. serve_sandwich: This action is for serving a sandwich to a child who is not allergic to gluten. The preconditions are that the child is not allergic to gluten, waiting at a place, and there is a sandwich on a tray at the same place. The effect is that the sandwich is no longer on the tray and the child is served.
6. move_tray: This action allows a tray to be moved from one place to another. The precondition is that the tray is at the starting place. The effect is that the tray is no longer at the starting place and is now at the destination place.
```
Q:
Domain PDDL:
```pddl
(define (domain termes)
(:requirements :typing :negative-preconditions)
(:types
    numb - object
    position - object
)
(:predicates
    (height ?p - position ?h - numb)
    (at ?p - position)
    (has-block)
    ;
    ; static predicates
    (SUCC ?n1 - numb ?n2 - numb)
    (NEIGHBOR ?p1 - position ?p2 - position)
    (IS-DEPOT ?p - position)
)
(:action move
    :parameters (?from - position ?to - position ?h - numb)
    :precondition
    (and
        (at ?from)
        (NEIGHBOR ?from ?to)
        (height ?from ?h)
        (height ?to ?h)
    )
    :effect
    (and
        (not (at ?from))
        (at ?to)
    )
)
(:action move-up
    :parameters (?from - position ?hfrom - numb ?to - position ?hto - numb)
    :precondition
    (and
        (at ?from)
        (NEIGHBOR ?from ?to)
        (height ?from ?hfrom)
        (height ?to ?hto)
        (SUCC ?hto ?hfrom)
    )
    :effect
    (and
        (not (at ?from))
        (at ?to)
    )
)
(:action move-down
    :parameters (?from - position ?hfrom - numb ?to - position ?hto - numb)
    :precondition
    (and
        (at ?from)
        (NEIGHBOR ?from ?to)
        (height ?from ?hfrom)
        (height ?to ?hto)
        (SUCC ?hfrom ?hto)
    )
    :effect
    (and
        (not (at ?from))
        (at ?to)
    )
)
(:action place-block
    :parameters (?rpos - position ?bpos - position ?hbefore - numb ?hafter - numb)
    :precondition
    (and
        (at ?rpos)
        (NEIGHBOR ?rpos ?bpos)
        (height ?rpos ?hbefore)
        (height ?bpos ?hbefore)
        (SUCC ?hafter ?hbefore)
        (has-block)
        (not (IS-DEPOT ?bpos))
    )
    :effect
    (and
        (not (height ?bpos ?hbefore))
        (height ?bpos ?hafter)
        (not (has-block))
    )
)
(:action remove-block
    :parameters (?rpos - position ?bpos - position ?hbefore - numb ?hafter - numb)
    :precondition
    (and
        (at ?rpos)
        (NEIGHBOR ?rpos ?bpos)
        (height ?rpos ?hafter)
        (height ?bpos ?hbefore)
        (SUCC ?hbefore ?hafter)
        (not (has-block))
    )
    :effect
    (and
        (not (height ?bpos ?hbefore))
        (height ?bpos ?hafter)
        (has-block)
    )
)
(:action create-block
    :parameters (?p - position)
    :precondition
    (and
        (at ?p)
        (not (has-block))
        (IS-DEPOT ?p)
    )
    :effect
    (and
        (has-block)
    )
)
(:action destroy-block
    :parameters (?p - position)
    :precondition
    (and
        (at ?p)
        (has-block)
        (IS-DEPOT ?p)
    )
    :effect
    (and
        (not (has-block))
    )
)
)
```
Problem PDDL:
```pddl
(define (problem prob)
(:domain termes)
; Initial state:
;  0   0  R0D
;  0   0   0
;  0   0   0
; Goal state:
;  0   0   0
;  0   1   0
;  0   0   0
; Maximal height: 1
(:objects
    n0 - numb
    n1 - numb
    pos-0-0 - position
    pos-0-1 - position
    pos-0-2 - position
    pos-1-0 - position
    pos-1-1 - position
    pos-1-2 - position
    pos-2-0 - position
    pos-2-1 - position
    pos-2-2 - position
)
(:init
    (height pos-0-0 n0)
    (height pos-0-1 n0)
    (height pos-0-2 n0)
    (height pos-1-0 n0)
    (height pos-1-1 n0)
    (height pos-1-2 n0)
    (height pos-2-0 n0)
    (height pos-2-1 n0)
    (height pos-2-2 n0)
    (at pos-2-0)
    (SUCC n1 n0)
    (NEIGHBOR pos-0-0 pos-1-0)
    (NEIGHBOR pos-0-0 pos-0-1)
    (NEIGHBOR pos-0-1 pos-1-1)
    (NEIGHBOR pos-0-1 pos-0-0)
    (NEIGHBOR pos-0-1 pos-0-2)
    (NEIGHBOR pos-0-2 pos-1-2)
    (NEIGHBOR pos-0-2 pos-0-1)
    (NEIGHBOR pos-1-0 pos-0-0)
    (NEIGHBOR pos-1-0 pos-2-0)
    (NEIGHBOR pos-1-0 pos-1-1)
    (NEIGHBOR pos-1-1 pos-0-1)
    (NEIGHBOR pos-1-1 pos-2-1)
    (NEIGHBOR pos-1-1 pos-1-0)
    (NEIGHBOR pos-1-1 pos-1-2)
    (NEIGHBOR pos-1-2 pos-0-2)
    (NEIGHBOR pos-1-2 pos-2-2)
    (NEIGHBOR pos-1-2 pos-1-1)
    (NEIGHBOR pos-2-0 pos-1-0)
    (NEIGHBOR pos-2-0 pos-2-1)
    (NEIGHBOR pos-2-1 pos-1-1)
    (NEIGHBOR pos-2-1 pos-2-0)
    (NEIGHBOR pos-2-1 pos-2-2)
    (NEIGHBOR pos-2-2 pos-1-2)
    (NEIGHBOR pos-2-2 pos-2-1)
    (IS-DEPOT pos-2-0)
)
(:goal
(and
    (height pos-0-0 n0)
    (height pos-0-1 n0)
    (height pos-0-2 n0)
    (height pos-1-0 n0)
    (height pos-1-1 n1)
    (height pos-1-2 n0)
    (height pos-2-0 n0)
    (height pos-2-1 n0)
    (height pos-2-2 n0)
    (not (has-block))
)
)
)
```
A:
```markdown
The Termes domain is a planning domain that simulates the behavior of robotic agents (inspired by termites) that can move around, pick up blocks, stack them to build structures, and remove blocks from structures. The domain includes actions for moving the robot, placing and removing blocks, and creating and destroying blocks at a depot.
The actions defined in this domain include:
1. move: This action allows the robot to move from one position to another at the same height. The preconditions are that the robot is at the starting position, the starting position is a neighbor to the destination position, and both positions have the same height. The effect is that the robot is no longer at the starting position and is now at the destination position.
2. move-up: This action allows the robot to move from a lower position to a neighboring higher position. The preconditions are that the robot is at the starting position, the starting position is a neighbor to the destination position, the starting position has a certain height, and the destination position's height is one less than the starting position's height. The effect is that the robot is no longer at the starting position and is now at the destination position.
3. move-down: This action allows the robot to move from a higher position to a neighboring lower position. The preconditions are that the robot is at the starting position, the starting position is a neighbor to the destination position, the starting position has a certain height, and the destination position's height is one less than the starting position's height. The effect is that the robot is no longer at the starting position and is now at the destination position.
4. place-block: This action allows the robot to place a block at a neighboring position, increasing the height of that position by one. The preconditions are that the robot is at a position next to the block position, both positions have the same height, the robot has a block, and the block position is not a depot. The effect is that the height of the block position is increased by one, and the robot no longer has a block.
5. remove-block: This action allows the robot to remove a block from a neighboring position, decreasing the height of that position by one. The preconditions are that the robot is at a position next to the block position, the robot's position is one height unit higher than the block position, and the robot does not have a block. The effect is that the height of the block position is decreased by one, and the robot now has a block.
6. create-block: This action allows the robot to create a block at a depot. The preconditions are that the robot is at the depot and does not have a block. The effect is that the robot now has a block.
7. destroy-block: This action allows the robot to destroy a block at a depot. The preconditions are that the robot is at the depot and has a block. The effect is that the robot no longer has a block.
```
Q:
Domain PDDL:
```pddl
(*@\highlb{[Target Domain PDDL Code]}@*)
```
Problem PDDL:
```pddl
(*@\highlb{[Target Problem PDDL Code]}@*)
```
\end{lstlisting}

%% file: figs/prompt_problem_translation.tex
\begin{lstlisting}[language=markdown, caption={Domain back-translation prompt template, with domain natural language description and problem PDDL placeholders for each target domain.}, label={lst:prompt_problem_translation}]
Your task is to translate problem PDDL files into natural language. Ensure that the resulting description covers all initial state and goal conditions. 
DO NOT be lazy in your response, be extremely precise in your descriptions such that all conditions are covered in your description and there is no ambiguity in your description.
If you do not find any common rule about some conditions, list all of them.
For the initial conditions, start with "Initially:", and for the goal conditions, start with "Your goal is to".
ALWAYS wrap your code in the appropriate markdown syntax.
Two examples are provided below.
Q:
Domain Description:
```markdown
The Termes domain is a planning domain that simulates the behavior of robotic agents (inspired by termites) that can move around, pick up blocks, stack them to build structures, and remove blocks from structures. The domain includes actions for moving the robot, placing and removing blocks, and creating and destroying blocks at a depot.
The actions defined in this domain include:
1. move: This action allows the robot to move from one position to another at the same height. The preconditions are that the robot is at the starting position, the starting position is a neighbor to the destination position, and both positions have the same height. The effect is that the robot is no longer at the starting position and is now at the destination position.
2. move-up: This action allows the robot to move from a lower position to a neighboring higher position. The preconditions are that the robot is at the starting position, the starting position is a neighbor to the destination position, the starting position has a certain height, and the destination position's height is one less than the starting position's height. The effect is that the robot is no longer at the starting position and is now at the destination position.
3. move-down: This action allows the robot to move from a higher position to a neighboring lower position. The preconditions are that the robot is at the starting position, the starting position is a neighbor to the destination position, the starting position has a certain height, and the destination position's height is one less than the starting position's height. The effect is that the robot is no longer at the starting position and is now at the destination position.
4. place-block: This action allows the robot to place a block at a neighboring position, increasing the height of that position by one. The preconditions are that the robot is at a position next to the block position, both positions have the same height, the robot has a block, and the block position is not a depot. The effect is that the height of the block position is increased by one, and the robot no longer has a block.
5. remove-block: This action allows the robot to remove a block from a neighboring position, decreasing the height of that position by one. The preconditions are that the robot is at a position next to the block position, the robot's position is one height unit higher than the block position, and the robot does not have a block. The effect is that the height of the block position is decreased by one, and the robot now has a block.
6. create-block: This action allows the robot to create a block at a depot. The preconditions are that the robot is at the depot and does not have a block. The effect is that the robot now has a block.
7. destroy-block: This action allows the robot to destroy a block at a depot. The preconditions are that the robot is at the depot and has a block. The effect is that the robot no longer has a block.
```
Problem PDDL:
```pddl
(define (problem prob)
(:domain termes)
; Initial state:
;  0   0  R0D
;  0   0   0
;  0   0   0
; Goal state:
;  0   0   0
;  0   1   0
;  0   0   0
; Maximal height: 1
(:objects
    n0 - numb
    n1 - numb
    pos-0-0 - position
    pos-0-1 - position
    pos-0-2 - position
    pos-1-0 - position
    pos-1-1 - position
    pos-1-2 - position
    pos-2-0 - position
    pos-2-1 - position
    pos-2-2 - position
)
(:init
    (height pos-0-0 n0)
    (height pos-0-1 n0)
    (height pos-0-2 n0)
    (height pos-1-0 n0)
    (height pos-1-1 n0)
    (height pos-1-2 n0)
    (height pos-2-0 n0)
    (height pos-2-1 n0)
    (height pos-2-2 n0)
    (at pos-2-0)
    (SUCC n1 n0)
    (NEIGHBOR pos-0-0 pos-1-0)
    (NEIGHBOR pos-0-0 pos-0-1)
    (NEIGHBOR pos-0-1 pos-1-1)
    (NEIGHBOR pos-0-1 pos-0-0)
    (NEIGHBOR pos-0-1 pos-0-2)
    (NEIGHBOR pos-0-2 pos-1-2)
    (NEIGHBOR pos-0-2 pos-0-1)
    (NEIGHBOR pos-1-0 pos-0-0)
    (NEIGHBOR pos-1-0 pos-2-0)
    (NEIGHBOR pos-1-0 pos-1-1)
    (NEIGHBOR pos-1-1 pos-0-1)
    (NEIGHBOR pos-1-1 pos-2-1)
    (NEIGHBOR pos-1-1 pos-1-0)
    (NEIGHBOR pos-1-1 pos-1-2)
    (NEIGHBOR pos-1-2 pos-0-2)
    (NEIGHBOR pos-1-2 pos-2-2)
    (NEIGHBOR pos-1-2 pos-1-1)
    (NEIGHBOR pos-2-0 pos-1-0)
    (NEIGHBOR pos-2-0 pos-2-1)
    (NEIGHBOR pos-2-1 pos-1-1)
    (NEIGHBOR pos-2-1 pos-2-0)
    (NEIGHBOR pos-2-1 pos-2-2)
    (NEIGHBOR pos-2-2 pos-1-2)
    (NEIGHBOR pos-2-2 pos-2-1)
    (IS-DEPOT pos-2-0)
)
(:goal
(and
    (height pos-0-0 n0)
    (height pos-0-1 n0)
    (height pos-0-2 n0)
    (height pos-1-0 n0)
    (height pos-1-1 n1)
    (height pos-1-2 n0)
    (height pos-2-0 n0)
    (height pos-2-1 n0)
    (height pos-2-2 n0)
    (not (has-block))
)
)
)
```
A:
```markdown
You control a robot capable of building structures by moving and manipulating blocks.
The environment consists of a grid of positions and two heights, numbered 0 and 1. The robot can move at the same height, move up one height, or move down one height. It can also place or remove a block at a neighboring position, or create or destroy a block at the depot. A block's height increases by one when placed and decreases by one when removed.
Initially:
- All positions on the grid have a height of 0.
- The robot is at position pos-2-0, which is the depot.
- The robot does not have a block.
- The positions have the following neighboring relationships:
  - pos-0-0 neighbors pos-1-0 and pos-0-1
  - pos-0-1 neighbors pos-1-1, pos-0-0, and pos-0-2
  - pos-0-2 neighbors pos-1-2 and pos-0-1
  - pos-1-0 neighbors pos-0-0, pos-2-0, and pos-1-1
  - pos-1-1 neighbors pos-0-1, pos-2-1, pos-1-0, and pos-1-2
  - pos-1-2 neighbors pos-0-2, pos-2-2, and pos-1-1
  - pos-2-0 neighbors pos-1-0 and pos-2-1, and is the depot
  - pos-2-1 neighbors pos-1-1, pos-2-0, and pos-2-2
  - pos-2-2 neighbors pos-1-2 and pos-2-1
- There is a successor relationship between the numbers n1 and n0.
Your goal is to achieve the following configuration:
- The height at pos-1-1 needs to be 1.
- All other positions must remain at height 0.
- The robot should not have a block at the end of the task.
```
Q:
Domain Description:
```markdown
The satellite domain is designed to model the operation of satellites that can take images of various targets in different modes. Each satellite is equipped with instruments that can be turned on and off, calibrated, and used to take images. The domain includes actions for turning the satellite to point at different directions, switching instruments on and off, calibrating instruments, and taking images.
The actions defined in this domain include:
1. turn_to: This action changes the direction the satellite is pointing. The preconditions are that the satellite must be pointing at a previous direction, and both the new and previous directions are valid. The effect is that the satellite is now pointing at the new direction and no longer pointing at the previous direction.
2. switch_on: This action turns on an instrument on board the satellite. The preconditions are that the instrument must be on board the satellite and there must be power available on the satellite. The effect is that the instrument is powered on, it is no longer calibrated, and the satellite no longer has power available.
3. switch_off: This action turns off an instrument on board the satellite. The preconditions are that the instrument must be on board the satellite and it must be powered on. The effect is that the satellite has power available and the instrument is no longer powered on.
4. calibrate: This action calibrates an instrument on board the satellite. The preconditions are that the satellite must be pointing at a calibration target for the instrument, the instrument must be on board the satellite and powered on. The effect is that the instrument is calibrated.
5. take_image: This action uses an instrument on board the satellite to take an image in a specific mode of a direction the satellite is pointing at. The preconditions are that the satellite must be pointing at the direction, the instrument must be calibrated, on board the satellite, support the mode, and be powered on. The effect is that an image of the direction in the specific mode is now available.
```
Problem PDDL:
```pddl
(define (problem strips-sat-x-1)
(:domain satellite)
(:objects
	satellite0
	instrument0
	satellite1
	instrument1
	instrument2
	instrument3
	satellite2
	instrument4
	instrument5
	instrument6
	satellite3
	instrument7
	satellite4
	instrument8
	thermograph2
	image3
	infrared1
	spectrograph4
	infrared0
	Star1
	Star4
	Star0
	GroundStation3
	Star2
	Star5
	Planet6
	Phenomenon7
	Star8
	Phenomenon9
	Star10
	Star11
	Star12
	Planet13
	Planet14
	Phenomenon15
	Planet16
	Star17
	Star18
	Planet19
)
(:init
	(satellite satellite0)
	(instrument instrument0)
	(supports instrument0 spectrograph4)
	(calibration_target instrument0 Star0)
	(on_board instrument0 satellite0)
	(power_avail satellite0)
	(pointing satellite0 Star8)
	(satellite satellite1)
	(instrument instrument1)
	(supports instrument1 infrared0)
	(supports instrument1 infrared1)
	(calibration_target instrument1 GroundStation3)
	(instrument instrument2)
	(supports instrument2 infrared1)
	(supports instrument2 infrared0)
	(calibration_target instrument2 Star2)
	(instrument instrument3)
	(supports instrument3 spectrograph4)
	(supports instrument3 infrared1)
	(supports instrument3 thermograph2)
	(calibration_target instrument3 Star0)
	(on_board instrument1 satellite1)
	(on_board instrument2 satellite1)
	(on_board instrument3 satellite1)
	(power_avail satellite1)
	(pointing satellite1 GroundStation3)
	(satellite satellite2)
	(instrument instrument4)
	(supports instrument4 infrared1)
	(supports instrument4 image3)
	(supports instrument4 infrared0)
	(calibration_target instrument4 Star2)
	(instrument instrument5)
	(supports instrument5 thermograph2)
	(supports instrument5 spectrograph4)
	(calibration_target instrument5 Star0)
	(instrument instrument6)
	(supports instrument6 infrared0)
	(calibration_target instrument6 GroundStation3)
	(on_board instrument4 satellite2)
	(on_board instrument5 satellite2)
	(on_board instrument6 satellite2)
	(power_avail satellite2)
	(pointing satellite2 Star4)
	(satellite satellite3)
	(instrument instrument7)
	(supports instrument7 image3)
	(calibration_target instrument7 Star2)
	(on_board instrument7 satellite3)
	(power_avail satellite3)
	(pointing satellite3 Phenomenon9)
	(satellite satellite4)
	(instrument instrument8)
	(supports instrument8 infrared0)
	(supports instrument8 spectrograph4)
	(supports instrument8 infrared1)
	(calibration_target instrument8 Star2)
	(on_board instrument8 satellite4)
	(power_avail satellite4)
	(pointing satellite4 Phenomenon9)
	(mode thermograph2)
	(mode image3)
	(mode infrared1)
	(mode spectrograph4)
	(mode infrared0)
	(direction Star1)
	(direction Star4)
	(direction Star0)
	(direction GroundStation3)
	(direction Star2)
	(direction Star5)
	(direction Planet6)
	(direction Phenomenon7)
	(direction Star8)
	(direction Phenomenon9)
	(direction Star10)
	(direction Star11)
	(direction Star12)
	(direction Planet13)
	(direction Planet14)
	(direction Phenomenon15)
	(direction Planet16)
	(direction Star17)
	(direction Star18)
	(direction Planet19)
)
(:goal (and
	(pointing satellite0 Phenomenon9)
	(pointing satellite1 Star4)
	(pointing satellite4 Star11)
	(have_image Star5 image3)
	(have_image Planet6 infrared1)
	(have_image Phenomenon7 infrared1)
	(have_image Star8 image3)
	(have_image Star10 thermograph2)
	(have_image Star11 infrared1)
	(have_image Planet13 spectrograph4)
	(have_image Planet14 thermograph2)
	(have_image Phenomenon15 infrared0)
	(have_image Planet16 image3)
	(have_image Star17 infrared0)
))
)
```
A:
```markdown
You are operating a constellation of satellites tasked with taking images of various targets in different modes.
Initially:
- There are five satellites (satellite0 to satellite4) with corresponding instruments (instrument0 to instrument8).
- Each instrument supports specific modes and has a calibration target:
  - Instrument0 supports spectrograph4 and targets Star0.
  - Instrument1 supports infrared0 and infrared1, targeting GroundStation3.
  - Instrument2 supports infrared1 and infrared0, targeting Star2.
  - Instrument3 supports spectrograph4, infrared1, and thermograph2, targeting Star0.
  - Instrument4 supports infrared1, image3, and infrared0, targeting Star2.
  - Instrument5 supports thermograph2 and spectrograph4, targeting Star0.
  - Instrument6 supports infrared0, targeting GroundStation3.
  - Instrument7 supports image3, targeting Star2.
  - Instrument8 supports infrared0, spectrograph4, and infrared1, targeting Star2.
- Instruments are on board their respective satellites, and all satellites have power available.
- Satellites are pointing at various directions:
  - Satellite0 is pointing at Star8.
  - Satellite1 is pointing at GroundStation3.
  - Satellite2 is pointing at Star4.
  - Satellite3 is pointing at Phenomenon9.
  - Satellite4 is pointing at Phenomenon9.
- There are various modes (thermograph2, image3, infrared1, spectrograph4, infrared0) and directions (Star1 to Star18, GroundStation3, Planet6, Phenomenon7, Phenomenon9, Planet13, Planet14, Phenomenon15, Planet16, Planet19).
Your goal is to:
- Point satellite0 at Phenomenon9.
- Point satellite1 at Star4.
- Point satellite4 at Star11.
- Have images of the following targets in the specified modes:
  - Star5 in image3 mode.
  - Planet6 in infrared1 mode.
  - Phenomenon7 in infrared1 mode.
  - Star8 in image3 mode.
  - Star10 in thermograph2 mode.
  - Star11 in infrared1 mode.
  - Planet13 in spectrograph4 mode.
  - Planet14 in thermograph2 mode.
  - Phenomenon15 in infrared0 mode.
  - Planet16 in image3 mode.
  - Star17 in infrared0 mode.
To achieve these goals, you will need to turn the satellites to point at the correct directions, switch on and calibrate the necessary instruments, and take images using the calibrated instruments in the supported modes.
```
Q:
Domain Description:
```markdown
(*@\highlb{[Target Domain Natural Language Description]}@*)
```
Problem PDDL:
```pddl
(*@\highlb{[Target Problem PDDL Code]}@*)
```
A:
\end{lstlisting}

%% file: figs/prompt_predicate_desc.tex
\begin{lstlisting}[language=markdown, caption={Predicate translation python code generation prompt.}, label={lst:prompt_predicate_translation}]
Your task is to generate python predicate descriptor for each environment. You are given the natural language description of the domain along with the PDDL code.
Q:
Domain Description:
```markdown
The robot has four actions: pickup, putdown, stack, and unstack. The domain assumes a world where there are a set of blocks that can be stacked on top of each other, an arm that can hold one block at a time, and a table where blocks can be placed.
The actions defined in this domain include:
pickup: allows the arm to pick up a block from the table if it is clear and the arm is empty. After the pickup action, the arm will be holding the block, and the block will no longer be on the table or clear.
putdown: allows the arm to put down a block on the table if it is holding a block. After the putdown action, the arm will be empty, and the block will be on the table and clear.
stack: allows the arm to stack a block on top of another block if the arm is holding the top block and the bottom block is clear. After the stack action, the arm will be empty, the top block will be on top of the bottom block, and the bottom block will no longer be clear.
unstack: allows the arm to unstack a block from on top of another block if the arm is empty and the top block is clear. After the unstack action, the arm will be holding the top block, the top block will no longer be on top of the bottom block, and the bottom block will be clear.
```
Domain PDDL:
```pddl
(define (domain blocksworld-4ops)
  (:requirements :strips)
(:predicates (clear ?x)
             (on-table ?x)
             (arm-empty)
             (holding ?x)
             (on ?x ?y))
(:action pickup
  :parameters (?ob)
  :precondition (and (clear ?ob) (on-table ?ob) (arm-empty))
  :effect (and (holding ?ob) (not (clear ?ob)) (not (on-table ?ob)) 
               (not (arm-empty))))
(:action putdown
  :parameters  (?ob)
  :precondition (holding ?ob)
  :effect (and (clear ?ob) (arm-empty) (on-table ?ob) 
               (not (holding ?ob))))
(:action stack
  :parameters  (?ob ?underob)
  :precondition (and (clear ?underob) (holding ?ob))
  :effect (and (arm-empty) (clear ?ob) (on ?ob ?underob)
               (not (clear ?underob)) (not (holding ?ob))))
(:action unstack
  :parameters  (?ob ?underob)
  :precondition (and (on ?ob ?underob) (clear ?ob) (arm-empty))
  :effect (and (holding ?ob) (clear ?underob)
               (not (on ?ob ?underob)) (not (clear ?ob)) (not (arm-empty)))))
```
A:
```python
def describe_predicate(predicate_name, predicate_args):
    """
    Predicates:
    - (clear ?x)
    - (on-table ?x)
    - (arm-empty)
    - (holding ?x)
    - (on ?x ?y)
    :param predicate_name: str
    :param predicate_args: Tuple[str]
    :return: Tuple[str, str] - (positive, negative)
    """
    # (clear ?x)
    if predicate_name == "clear":
        (x,) = predicate_args
        return f"Block {x} is clear.", f"Block {x} is not clear."
    # (on-table ?x)
    elif predicate_name == "on-table":
        (x,) = predicate_args
        return f"Block {x} is on the table.", f"Block {x} is not on the table."
    # (arm-empty)
    elif predicate_name == "arm-empty":
        return "Arm is empty.", "Arm is not empty."
    # (holding ?x)
    elif predicate_name == "holding":
        (x,) = predicate_args
        return f"Arm is holding block {x}.", f"Arm is not holding block {x}."
    # (on ?x ?y)
    elif predicate_name == "on":
        (x, y) = predicate_args
        return f"Block {x} is on block {y}.", f"Block {x} is not on block {y}."
    else:
        raise ValueError(f"Unknown predicate: {predicate_name}")
```
Q:
Domain Description:
```markdown
(*@\highlb{[Target Domain Natural Language Description]}@*)
```
Domain PDDL:
```pddl
(*@\highlb{[Target Domain PDDL Code]}@*)
```
A:

\end{lstlisting}